\documentclass[runningheads]{llncs}
\usepackage{graphicx}
\usepackage{color}
\usepackage{graphicx}
\usepackage{amsmath,amssymb} 
\usepackage{epsfig}
\usepackage{amsmath}
\usepackage{amssymb}
\usepackage{subfigure}
\usepackage{epsfig}
\usepackage{multirow}
\usepackage{epstopdf}
\usepackage{graphics}
\usepackage{graphicx}
\usepackage{amsmath}
\usepackage{amssymb}
\usepackage{wrapfig}
\usepackage{enumerate}
\usepackage[breaklinks=true,letterpaper=true,colorlinks,bookmarks=false]{hyperref}
\usepackage[width=122mm,left=12mm,paperwidth=146mm,height=193mm,top=12mm,paperheight=217mm]{geometry}
\begin{document}

\def\ECCV14SubNumber{690}  

\title{Multi-Scale Orderless Pooling of\\Deep Convolutional Activation Features} 

\titlerunning{Multi-scale Orderless Pooling of Deep Convolutional Activation Features}

\authorrunning{Y. Gong et al.}

\author{Yunchao Gong$^1$,~ Liwei Wang$^2$,~ Ruiqi Guo$^2$,~ and~ Svetlana Lazebnik$^2$}
\institute{$^1$University of North Carolina at Chapel Hill\\
{\tt yunchao@cs.unc.edu}\\
$^2$University of Illinois at Urbana-Champaign\\
{\tt \{lwang97,guo29,slazebni\}@illinois.edu}}

\addtolength{\textfloatsep}{-5mm}
\addtolength{\floatsep}{-5mm}

\maketitle

\begin{abstract}
Deep convolutional neural networks (CNN) have shown their promise as a universal representation for recognition. However, global CNN activations lack geometric invariance, which limits their robustness for classification and matching of highly variable scenes. To improve the invariance of CNN activations without degrading their discriminative power, this paper presents a simple but effective scheme called {\em multi-scale orderless pooling} (MOP-CNN). This scheme extracts CNN activations for local patches at multiple scale levels, performs orderless VLAD pooling of these activations at each level separately, and concatenates the result. The resulting MOP-CNN representation can be used as a generic feature for either supervised or unsupervised recognition tasks, from image classification to instance-level retrieval; it consistently outperforms global CNN activations without requiring any joint training of prediction layers for a particular target dataset. In absolute terms, it achieves state-of-the-art results on the challenging SUN397 and MIT Indoor Scenes classification datasets, and competitive results on ILSVRC2012/2013 classification and INRIA Holidays retrieval datasets.
\end{abstract}

\section{Introduction}

Recently, deep convolutional neural networks (CNN)~\cite{lecun90} have demonstrated breakthrough accuracies for image classification~\cite{krizhevsky2012imagenet}. This has spurred a flurry of activity on further improving CNN architectures and training algorithms \cite{goodfellow13,le38115,wan13,Hinton12,simonyan13deep}, as well as on using CNN features as a universal representation for recognition. A number of recent works~\cite{donahue2013decaf,girshick2013rich,oquab2014learning,Razavian14,sermanet2013overfeat} show that CNN features trained on sufficiently large and diverse datasets such as ImageNet~\cite{deng09} can be successfully transferred to other visual recognition tasks, e.g., scene classification and object localization, with a only limited amount of task-specific training data. Our work also relies on reusing CNN activations as off-the-shelf features for whole-image tasks like scene classification and retrieval. But, instead of simply computing the CNN activation vector over the entire image, we ask whether we can get improved performance by combining activations extracted at multiple {\em local} image windows. Inspired by previous work on spatial and feature space pooling of local descriptors~\cite{lazebnik06,Perronnin07,jegou2010aggregating}, we propose a novel and simple pooling scheme that significantly outperforms global CNN activations for both supervised tasks like image classification and unsupervised tasks like retrieval, even without any fine-tuning on the target datasets.

Image representation has been a driving motivation for research in computer vision for many years. For much of the past decade, orderless bag-of-features (BoF) methods~\cite{Perronnin07,wangjj10,Csurka04,Sivic03,Grauman05thepyramid} were considered to be the state of the art. Especially when built on top of locally invariant features like SIFT~\cite{lowe2004distinctive}, BoF can be, to some extent, robust to image scaling, translation, occlusion, and so on. However, they do not encode global spatial information, motivating the incorporation of loose spatial information in the BoF vectors through spatial pyramid matching (SPM)~\cite{lazebnik06}. Deep CNN, as exemplified by the system of Krizhevsky et al.~\cite{krizhevsky2012imagenet}, is a completely different architecture. Raw image pixels are first sent through five convolutional layers, each of which filters the feature maps and then max-pools the output within local neighborhoods. At this point, the representation still preserves a great deal of global spatial information. For example, as shown by Zeiler and Fergus~\cite{zeiler2013visualizing}, the activations from the fifth max-pooling layer can be reconstructed to form an image that looks similar to the original one. Though max-pooling within each feature map helps to improve invariance to small-scale deformations~\cite{lee2009convolutional}, invariance to larger-scale, more global deformations might be undermined by the preserved spatial information. After the filtering and max-pooling layers follow several fully connected layers, finally producing an activation of 4096 dimensions. While it becomes more difficult to reason about the invariance properties of the output of the fully connected layers, we will present an empirical analysis in Section \ref{s3} indicating that the final CNN representation is still fairly sensitive to global translation, rotation, and scaling. Even if one does not care about this lack of invariance for its own sake, we show that it directly translates into a loss of accuracy for classification tasks.

\begin{figure}[t]
   \hspace{-1.1cm}
   \includegraphics[width=6.8in, clip=true, trim= 0mm 2mm -50mm 0mm]{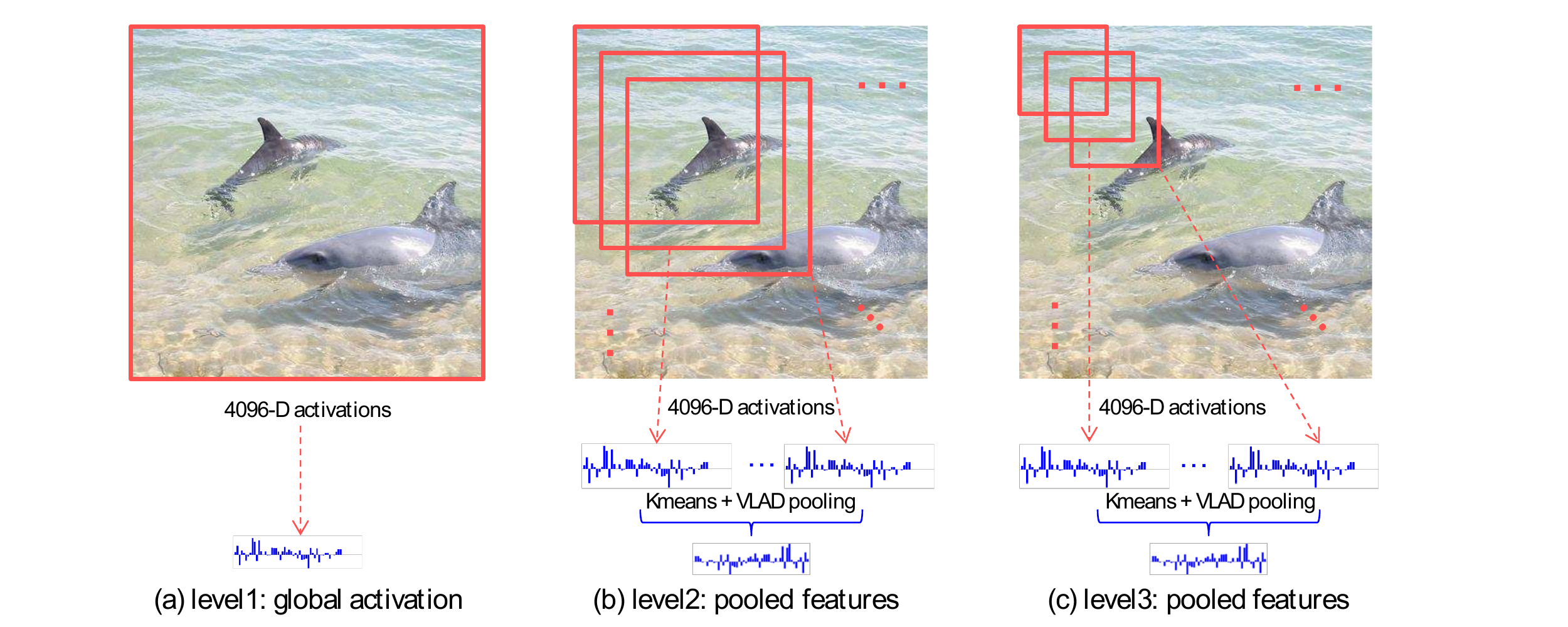}
   \caption{Overview of multi-scale orderless pooling for CNN activations (MOP-CNN). Our proposed feature is a concatenation of the feature vectors from three levels: (a) Level 1, corresponding to the 4096-dimensional CNN activation for the entire $256\times256$ image; (b) Level 2, formed by extracting activations from $128\times128$ patches and VLAD pooling them with a codebook of 100 centers; (c) Level 3, formed in the same way as level 2 but with $64\times64$ patches.
   \label{idea}}
\end{figure}

Intuitively, bags of features and deep CNN activations lie towards opposite ends of the ``orderless'' to ``globally ordered'' spectrum for visual representations. SPM~\cite{lazebnik06} is based on realizing that BoF has insufficient spatial information for many recognition tasks and adding just enough such information. Inspired by this, we observe that CNN activations preserve too much spatial information, and study the question of whether we can build a more orderless representation on top of CNN activations to improve recognition performance. We present a simple but effective framework for doing this, which we refer to as {\em multi-scale orderless pooling} (MOP-CNN). The idea is summarized in Figure \ref{idea}. Briefly, we begin by extracting deep activation features from local patches at multiple scales. Our coarsest scale is the whole image, so global spatial layout is still preserved, and our finer scales allow us to capture more local, fine-grained details of the image. Then we aggregate local patch responses at the finer scales via VLAD encoding \cite{jegou2010aggregating}. The orderless nature of VLAD helps to build a more invariant representation. Finally, we concatenatenate the original global deep activations with the VLAD features for the finer scales to form our new image representation.

Section \ref{s4} will introduce our multi-scale orderless pooling approach. Section \ref{s3} will present a small-scale study suggesting that CNN activations extracted at sub-image windows can provide more robust and discriminative information than whole-image activations, and confirming that MOP-CNN is more robust in the presence of geometric deformations than global CNN. Next, Section \ref{experiments} will report comprehensive experiments results for classification on three image datasets (SUN397, MIT Indoor Scenes, and ILSVRC2012/2013) and retrieval on the Holidays dataset. A sizable boost in performance across these popular benchmarks confirms the promise of our method. Section \ref{sec:discussion} will conclude with a discussion of future work directions.


\section{The Proposed Method \label{s4}}

Inspired by SPM~\cite{lazebnik06}, which extracts local patches at a single scale but then pools them over regions of increasing scale, ending with the whole image, we propose a kind of ``reverse SPM'' idea, where we extract patches at multiple scales, starting with the whole image, and then pool each scale without regard to spatial information. The basic idea is illustrated in Figure \ref{idea}.

Our representation has three scale levels, corresponding to CNN activations of the global $256\times256$ image and $128\times 128$ and $64 \times 64$ patches, respectively. To extract these activations, we use the Caffe CPU implementation~\cite{Jia13caffe} pre-trained on ImageNet~\cite{deng09}. Given an input image or a patch, we resample it to $256\times256$ pixels, subtract the mean of the pixel values, and feed the patch through the network. Then we take the 4096-dimensional output of the seventh (fully connected) layer, after the rectified linear unit (ReLU) transformation, so that all the values are non-negative (we have also tested the activations before ReLU and found worse performance).

For the first level, we simply take the 4096-dimensional CNN activation for the whole $256\times256$ image.
For the remaining two levels, we extract activations for all $128\times 128$ and $64 \times 64$ patches sampled with a stride of 32 pixels. Next, we need to pool the activations of these multiple patches to summarize the second and third levels by single feature vectors of reasonable dimensionality. For this, we adopt Vectors of Locally Aggregated Descriptors (VLAD)~\cite{jegou2010aggregating,Bergamo:2013}, which are a simplified version of Fisher Vectors (FV)~\cite{Perronnin07}. 
At each level, we extract the 4096-dimensional activations for respective patches and, to make computation more efficient, use PCA to reduce them to 500 dimensions. We also learn a separate $k$-means codebook for each level with $k=100$ centers.
Given a collection of patches from an input image and a codebook of centers $\boldsymbol c_i,~i=1,\ldots,k$,
the VLAD descriptor (soft assignment version from \cite{Bergamo:2013}) is constructed by assigning each patch $\boldsymbol p_j$ to its $r$ nearest cluster centers $r$NN$(\boldsymbol p_j)$ 
and aggregating the residuals of the patches minus the center:
\begin{equation*}\small
  \boldsymbol x = \left[\sum_{j:\,\boldsymbol c_1 \in r\mathrm{NN}(\boldsymbol p_j)} w_{j1} (\boldsymbol p_j - \boldsymbol c_1) , ~~\ldots~~, \sum_{j:\,\boldsymbol c_{k} \in r\mathrm{NN}(\boldsymbol p_j) } w_{jk}(\boldsymbol p_j - \boldsymbol c_{k}) \right] \,,
\end{equation*}\normalsize
where $w_{jk}$ is the Gaussian kernel similarity between $\boldsymbol p_j$ and $\boldsymbol c_k$. For each patch, we additionally normalize its weights to its nearest $r$ centers to have sum one. For the results reported in the paper, we use $r=5$\footnote{In the camera-ready version of the paper, we incorrectly reported using $r=1$, which is equivalent to the hard assignment VLAD in \cite{jegou2010aggregating}. However, we have experimented with different $r$ and their accuracy on our datasets is within 1\% of each other.} and kernel standard deviation of 10. Following \cite{jegou2010aggregating}, we power- and  L2-normalize the pooled vectors. However, the resulting vectors still have quite high dimensionality: given 500-dimensional patch activations $\boldsymbol p_j$ (after PCA) and 100 $k$-means centers, we end up with 50,000 dimensions. This is too high for many large-scale applications, so we further perform PCA on the pooled vectors and reduce them to 4096 dimensions. Note that applying PCA after the two stages (local patch activation and global pooled vector) is a standard practice in previous works~\cite{Perronnin10eccv,Perronnin10cvpr}. Finally, given the original 4096-dimensional feature vector from level one and the two 4096-dimensional pooled PCA-reduced vectors from levels two and three, we rescale them to unit norm and concatenate them to form our final image representation.
\smallskip

\section{Analysis of Invariance \label{s3}}

\begin{figure*}[bpt]
\centering
  \includegraphics[width= 1\linewidth]{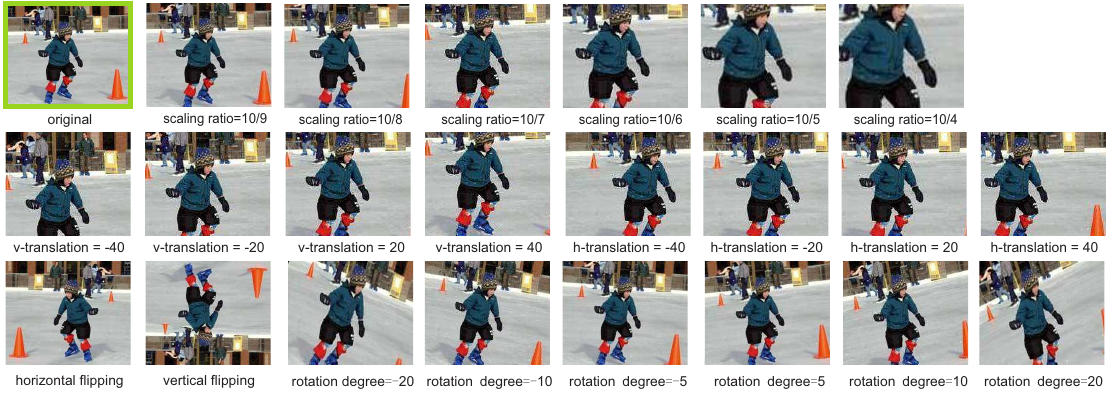} \vspace{-6mm}
    \caption{Illustration of image transformations considered in our invariance study. For scaling by a factor of $\rho$, we take crops around the image center of $(1/\rho)$ times original size. For translation, we take crops of 0.7 times the original size and translate them by up to 40 pixels in either direction horizontally or vertically (the translation amount is relative to the normalized image size of $256\times 256$). For rotation, we take crops from the middle of the image (so as to avoid corner artifacts) and rotate them from -20 to 20 degrees about the center. The corresponding scaling ratio, translation distance (pixels) and rotation degrees are listed below each instance.}\label{fig:trans}
\end{figure*}

\begin{figure*}[!h]
\begin{center}
\begin{tabular}{l}
{\includegraphics[width=1.3in]{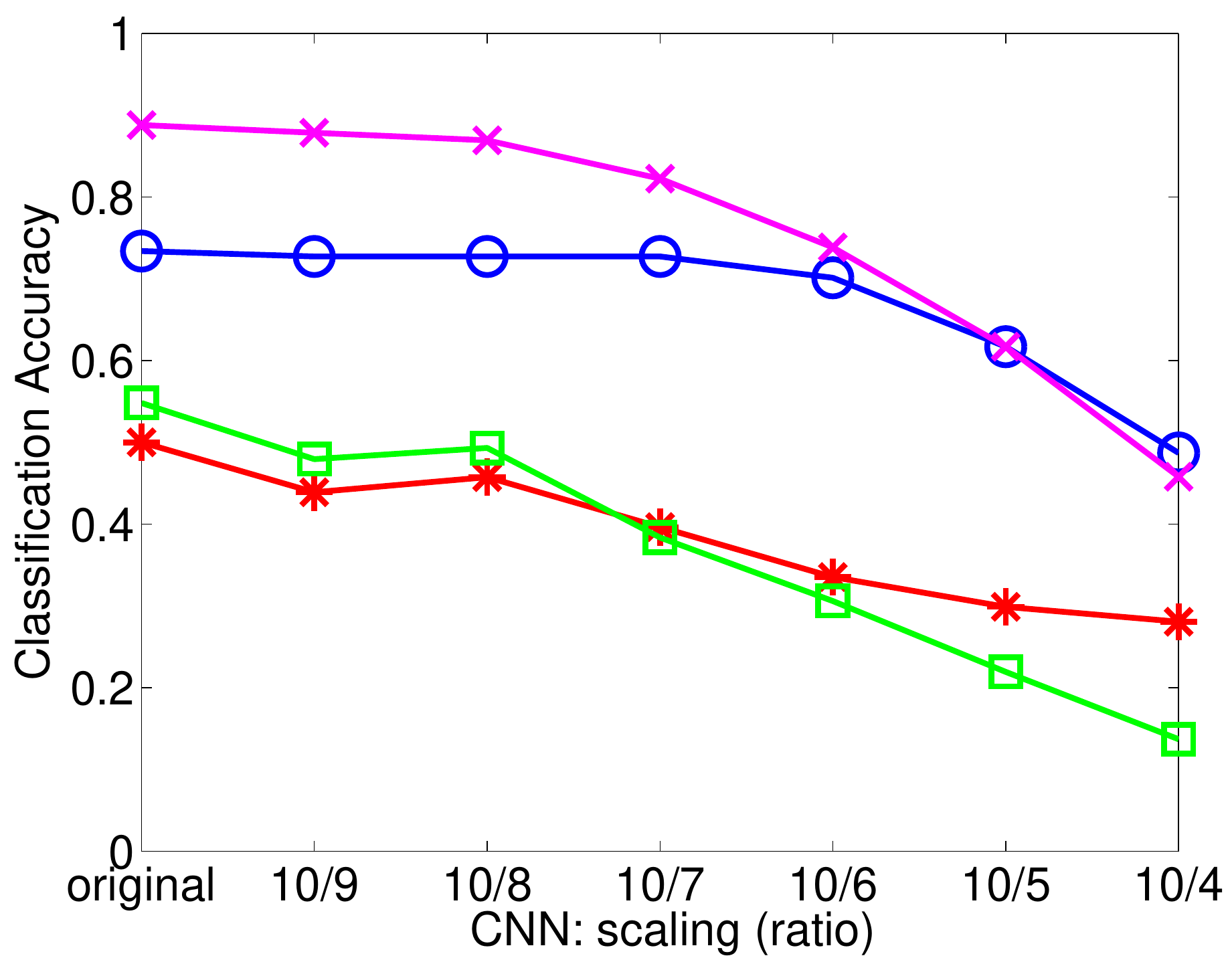}}
{\includegraphics[width=1.3in]{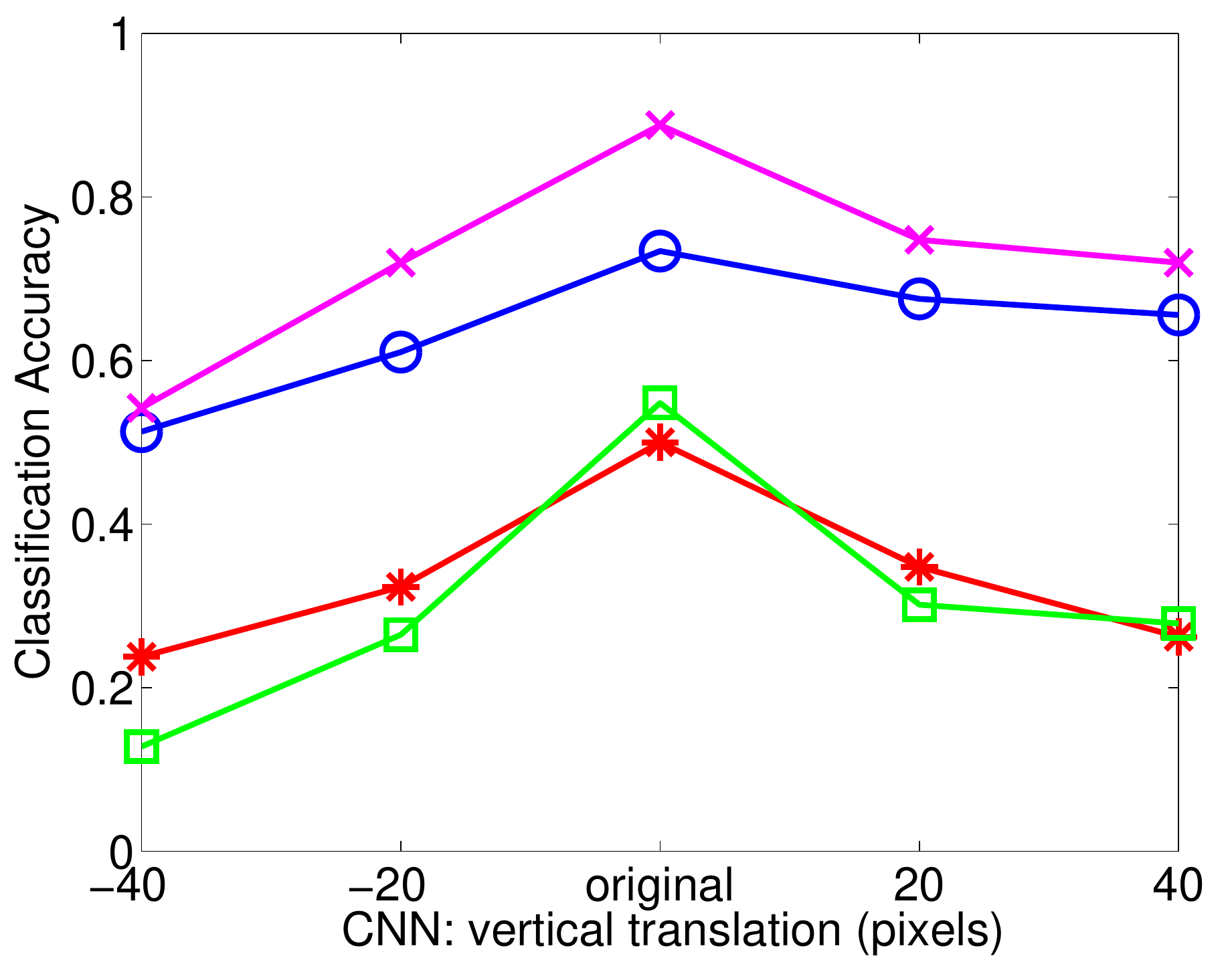}}
{\includegraphics[width=1.3in]{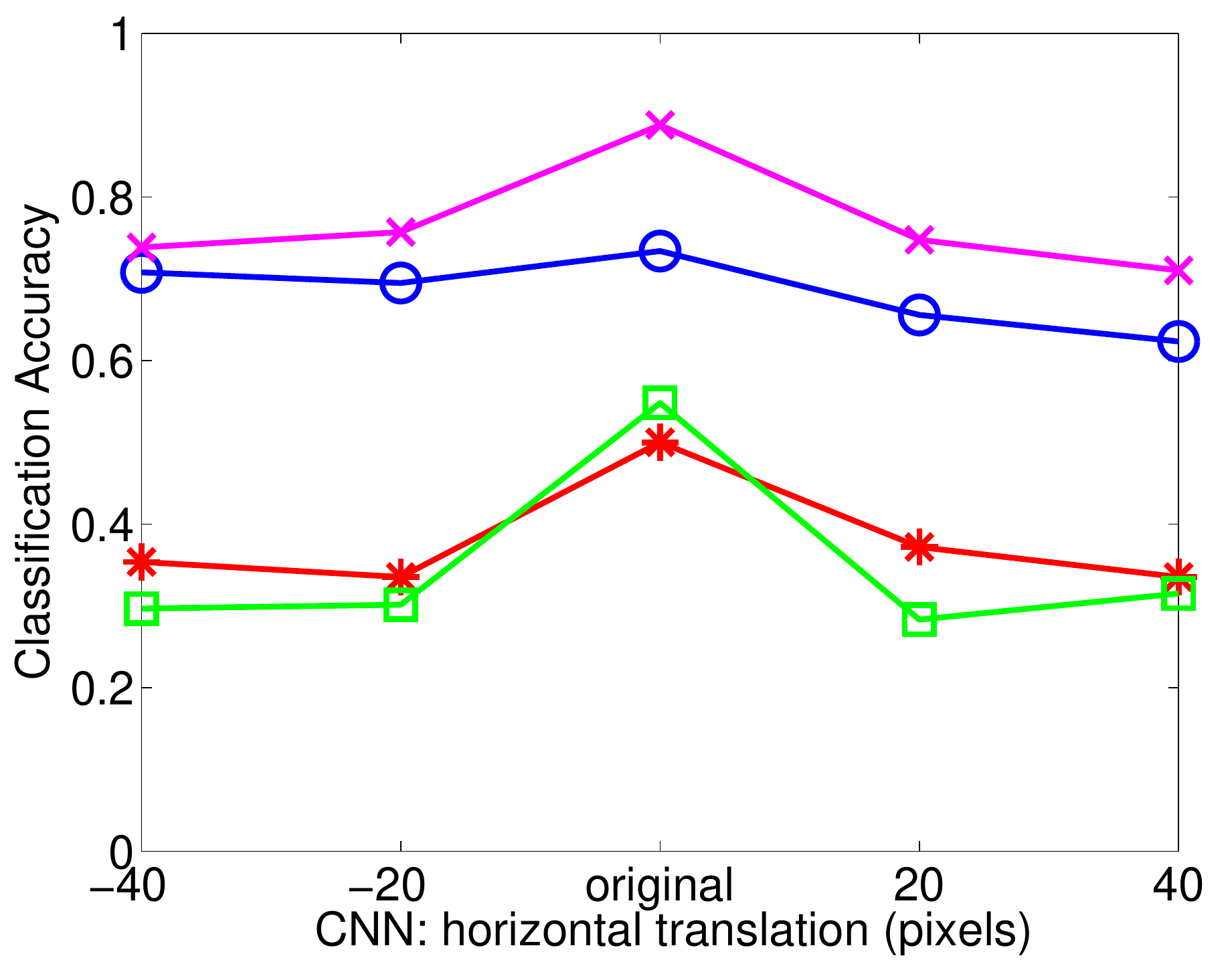}}\\
\subfigure[scaling]{\includegraphics[width=1.3in]{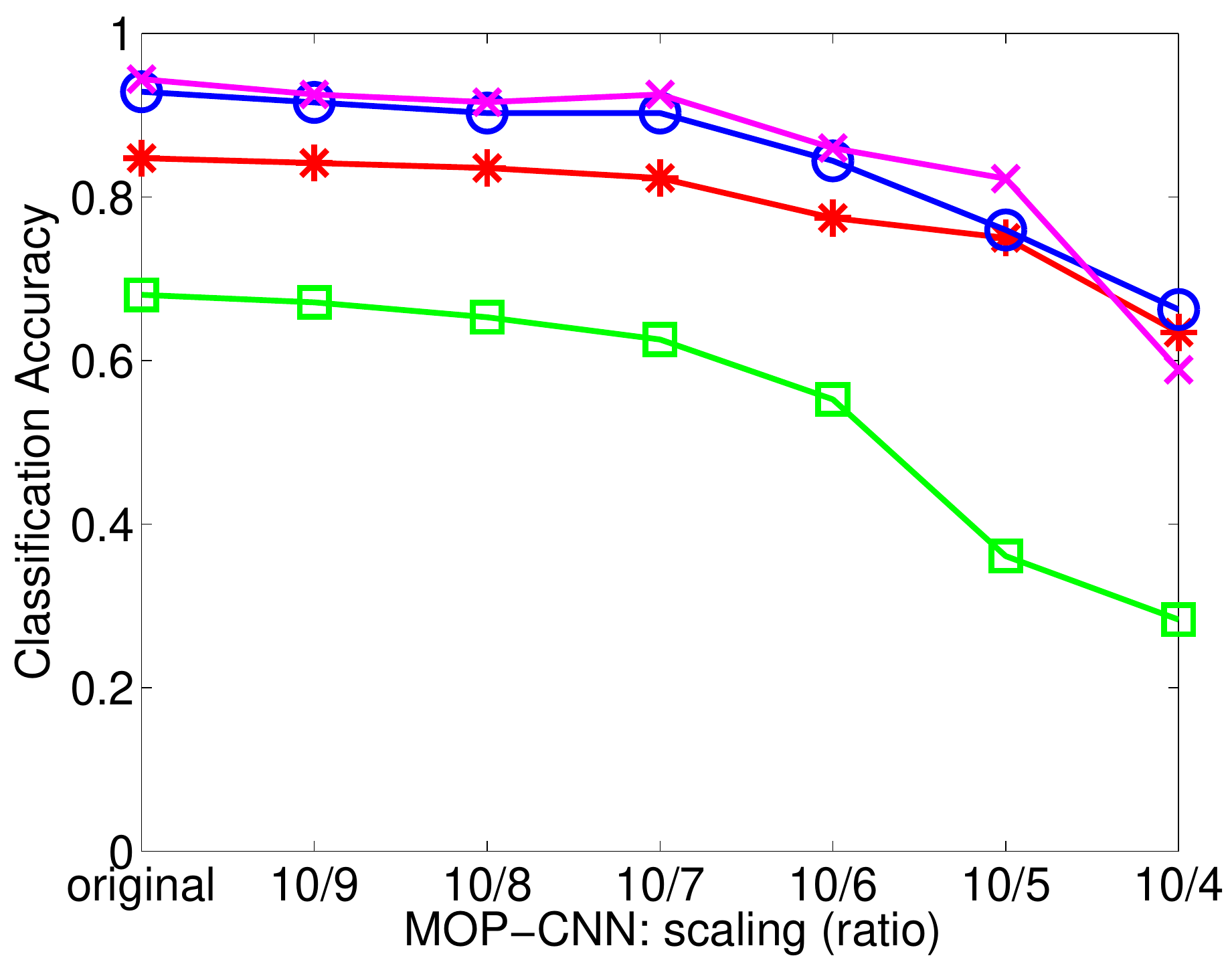}}
\subfigure[v-translation]{\includegraphics[width=1.3in]{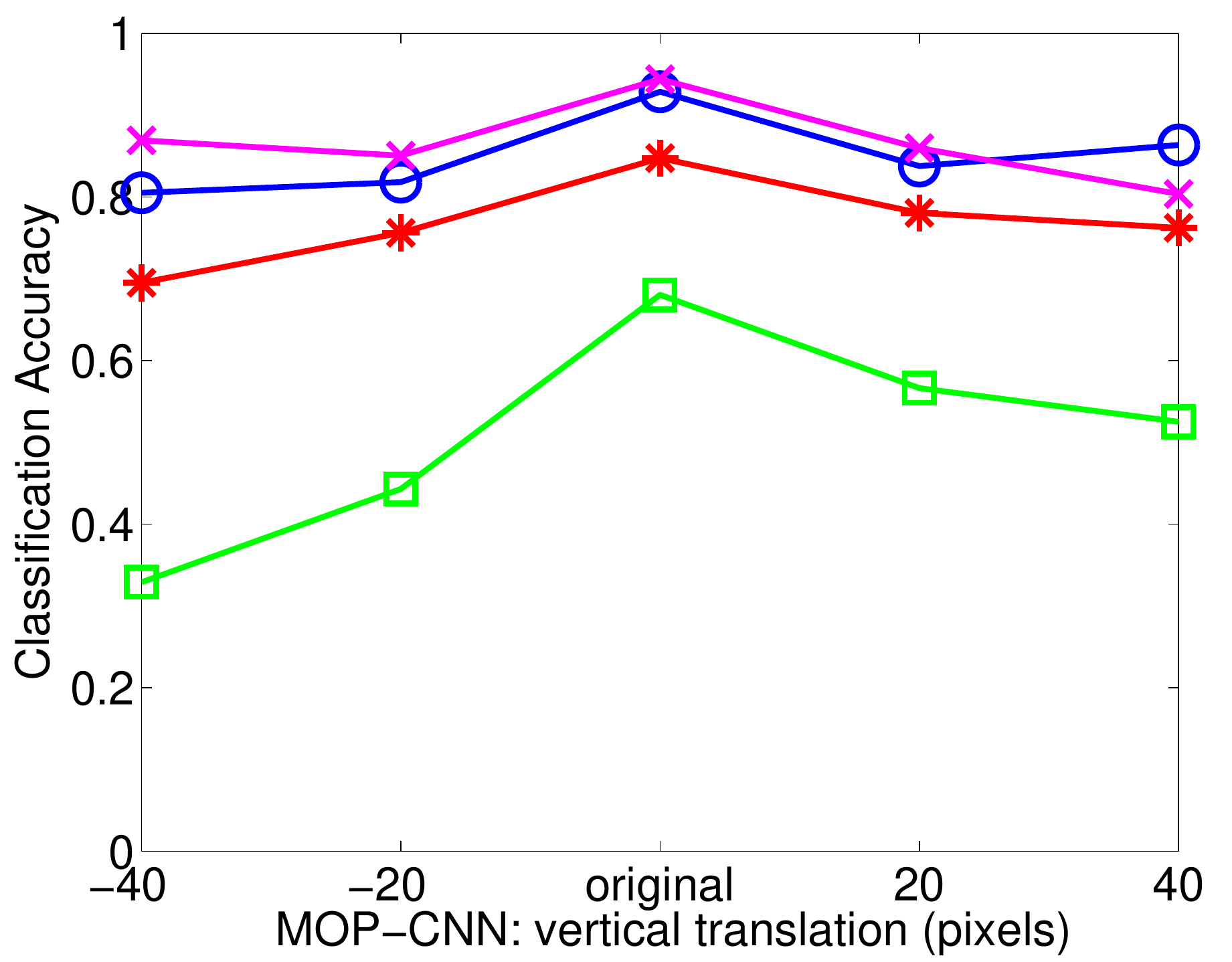}}
\subfigure[h-translation]{\includegraphics[width=1.3in]{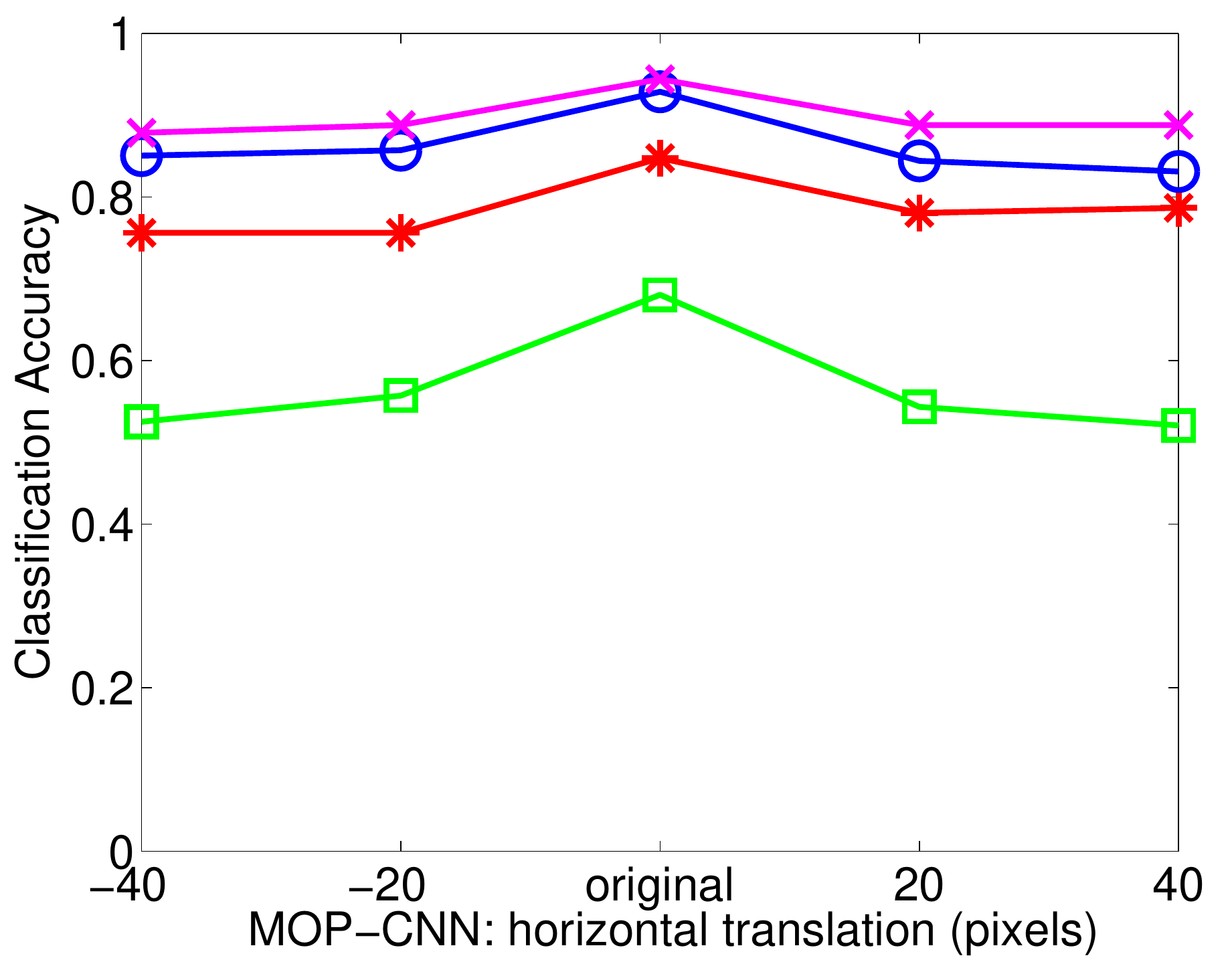}}\\
{\includegraphics[width=1.3in]{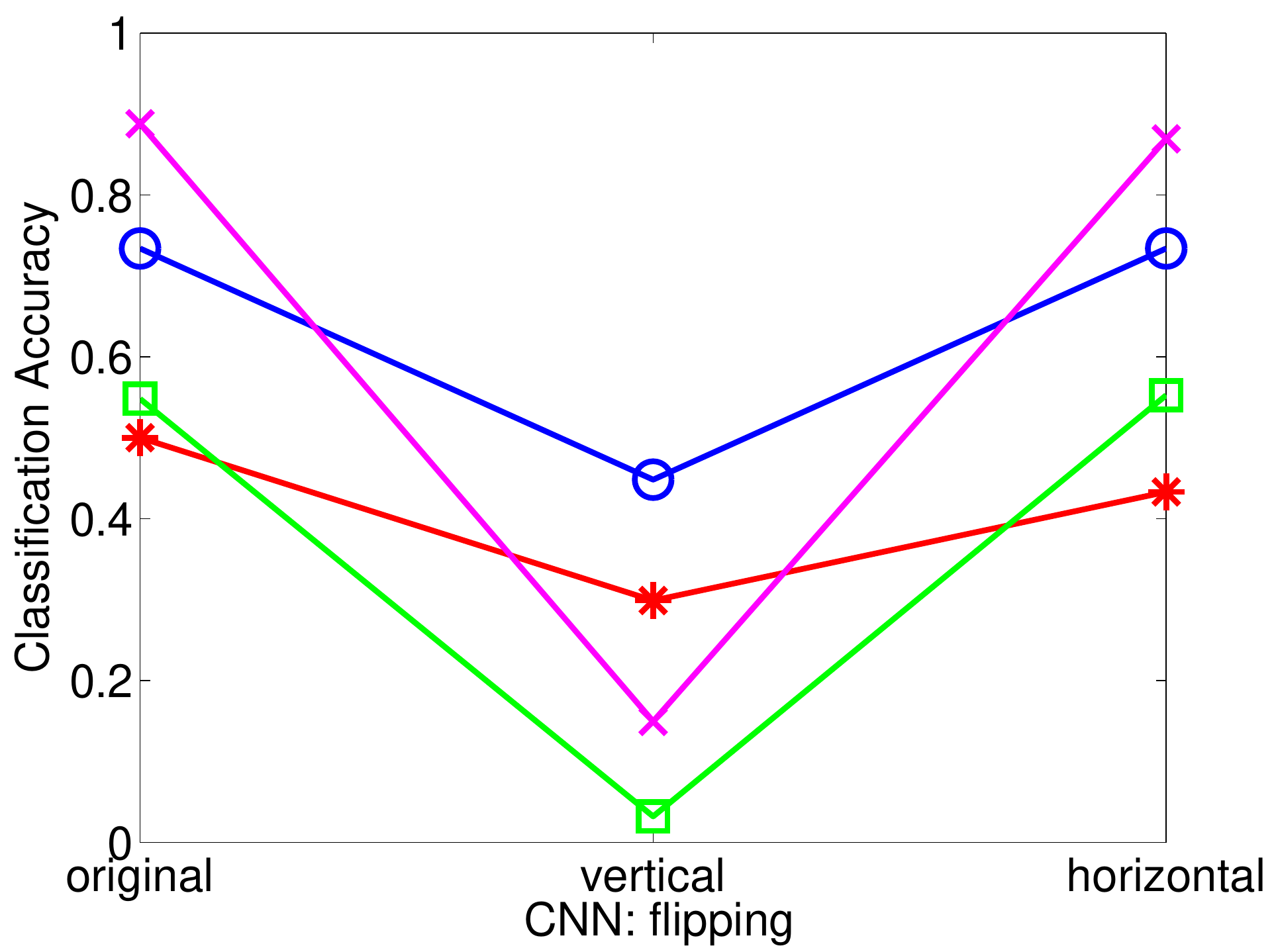}}
{\includegraphics[width=1.3in]{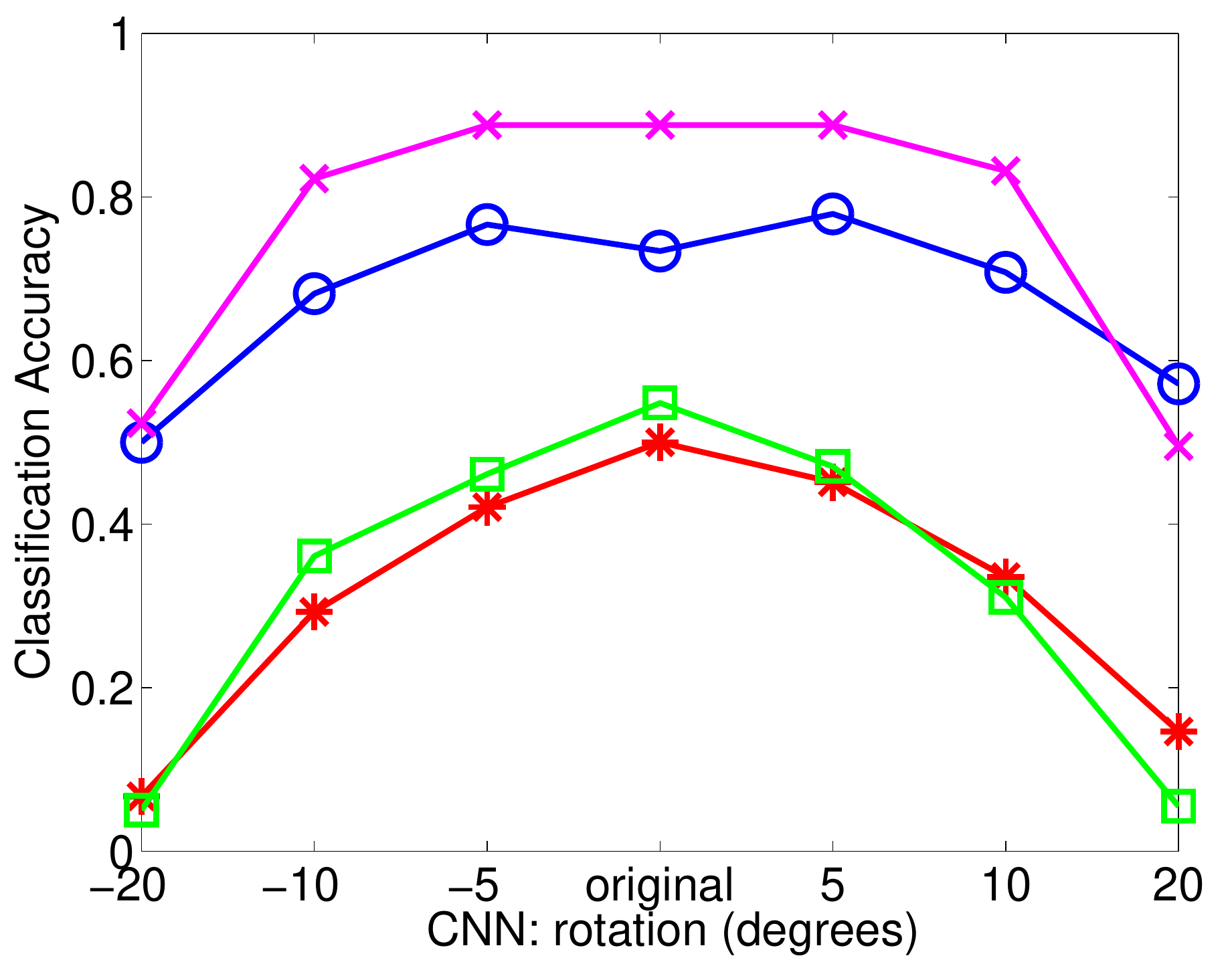}}\\
\subfigure[flipping]{\includegraphics[width=1.3in]{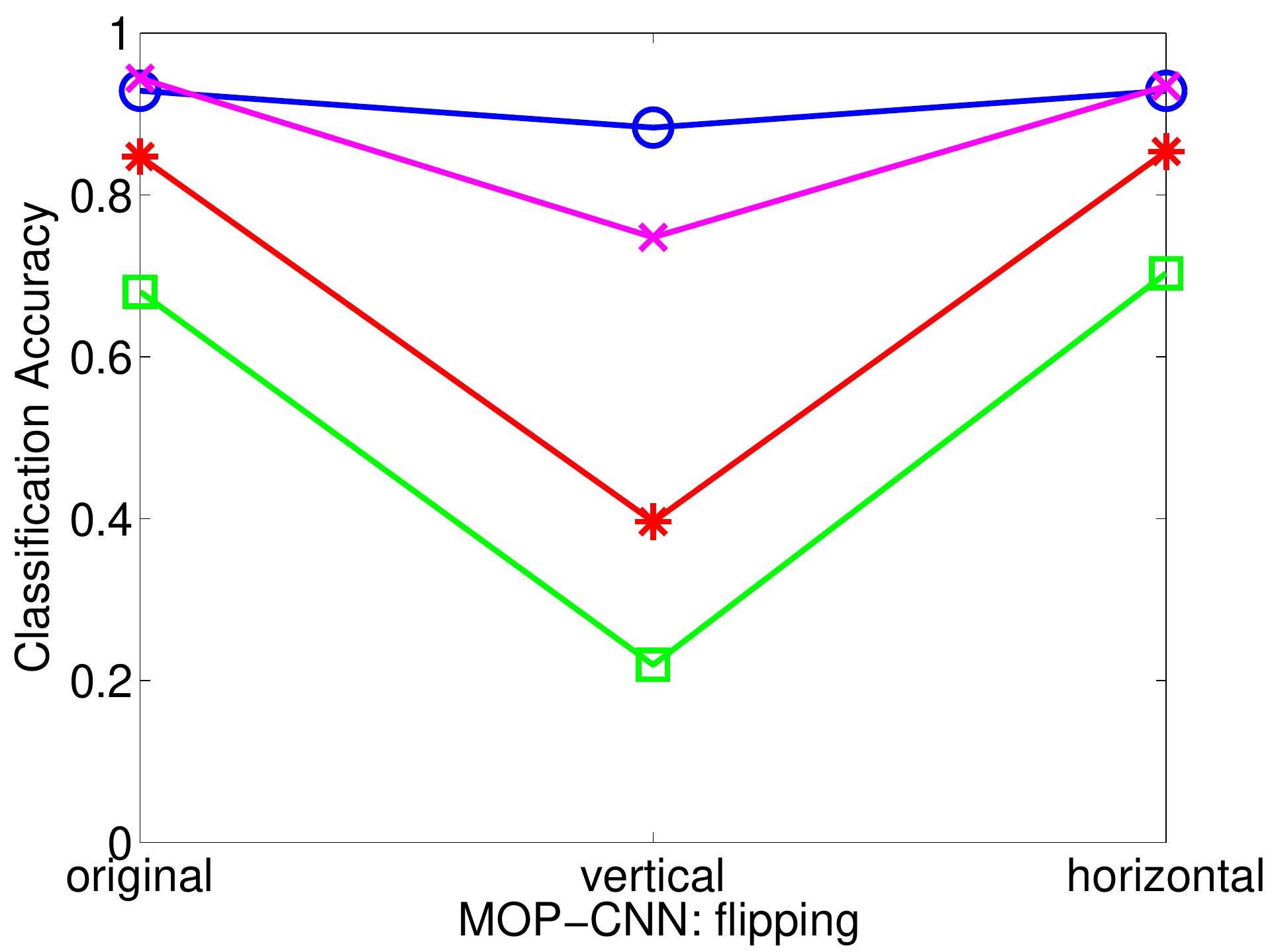}}
\subfigure[rotation]{\includegraphics[width=1.3in]{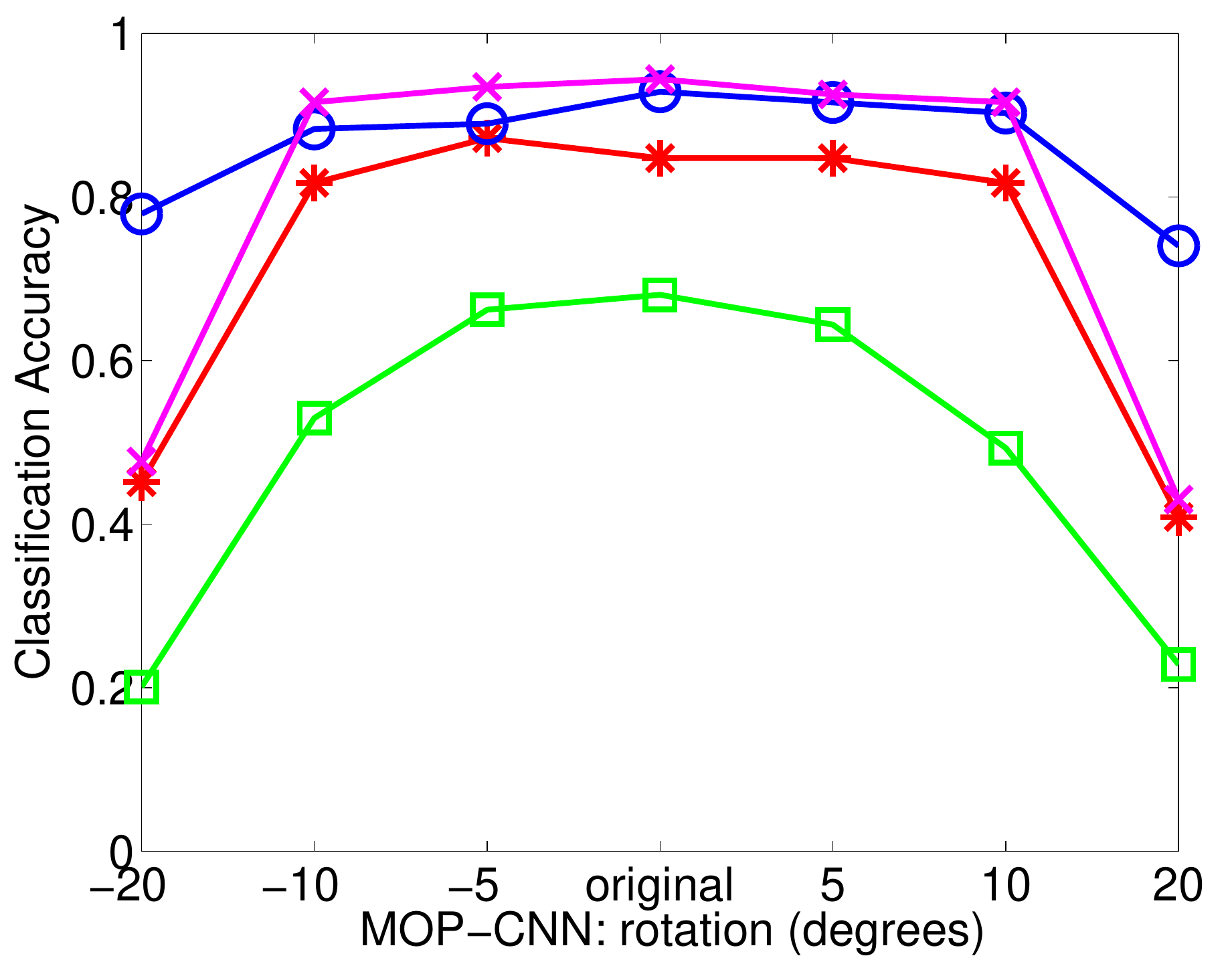}}
\subfigure{\includegraphics[width=1in,trim= -20mm -60mm 20mm 60mm]{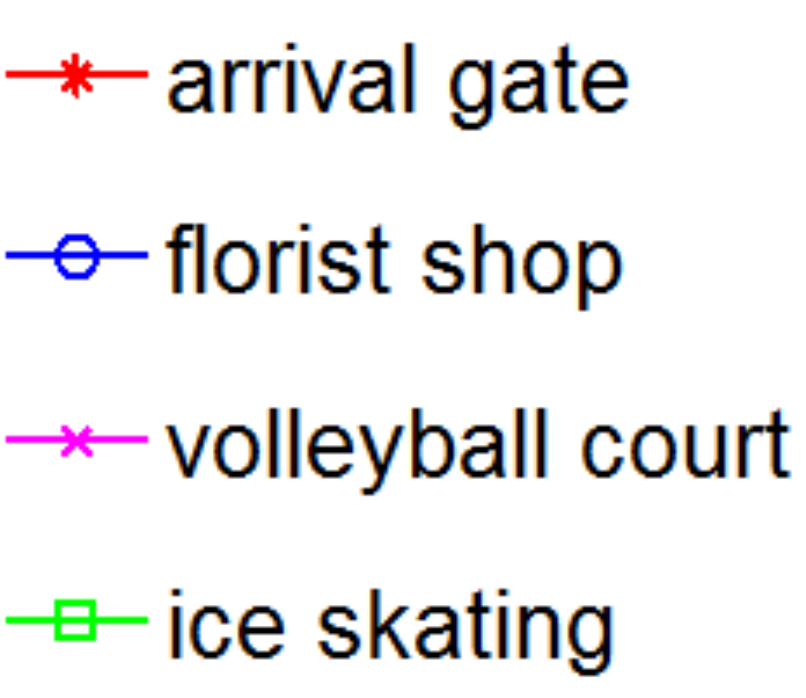}}
\end{tabular} \vspace{-2mm}
\caption{Accuracies for 397-way classification on four classes from the SUN dataset as a function of different transformations of the test images. For each transformation type (a-e), the upper (resp. lower) plot shows the classification accuracy using the global CNN representation (resp. MOP-CNN).}\vspace{-5mm}
\label{fig:statistics}
\end{center}
\end{figure*}

We first examine the invariance properties of global CNN activations vs. MOP-CNN. As part of their paper on visualizing deep features, Zeiler and Fergus~\cite{zeiler2013visualizing} analyze the transformation invariance of their model on five individual images by displaying the distance between the feature vectors of the original and transformed images, as well as the change in the probability of the correct label for the transformed version of the image (Figure 5 of~\cite{zeiler2013visualizing}). These plots show very different patterns for different images, making it difficult to draw general conclusions. We would like to conduct a more comprehensive analysis with an emphasis on prediction accuracy for entire categories, not just individual images. To this end, we train one-vs-all linear SVMs on the original training images for all 397 categories from the SUN dataset~\cite{xiao2010sun} using both global 4096-dimensional CNN activations and our proposed MOP-CNN features. 
At test time, we consider four possible transformations: translation, scaling, flipping and rotation (see Figure \ref{fig:trans} for illustration and detailed explanation of transformation parameters). We apply a given transformation to all the test images, extract features from the transformed images, and perform 397-way classification using the trained SVMs. Figure \ref{fig:statistics} shows classification accuracies as a function of transformation type and parameters for four randomly selected classes: arrival gate, florist shop, volleyball court, and ice skating.
In the case of CNN features, for almost all transformations, as the degree of transformation becomes more extreme, the classification accuracies keep dropping for all classes. The only exception is horizontal flipping, which does not seem to affect the classification accuracy. This may be due to the fact that the Caffe implementation adds horizontal flips of all training images to the training set (on the other hand, the Caffe training protocol also involves taking random crops of training images, yet this does not seem sufficient for building in invariance to such transformations, as our results indicate). By contrast with global CNN, our MOP-CNN features are more robust to the degree of translation, rotation, and scaling, and their absolute classification accuracies are consistently higher as well.

\begin{figure*}[t]
\hspace{-0.4cm}
\begin{center}
\begin{tabular}{c}
\subfigure[ski]{\includegraphics[height=0.75in,trim=0mm 17mm 0mm 5mm]{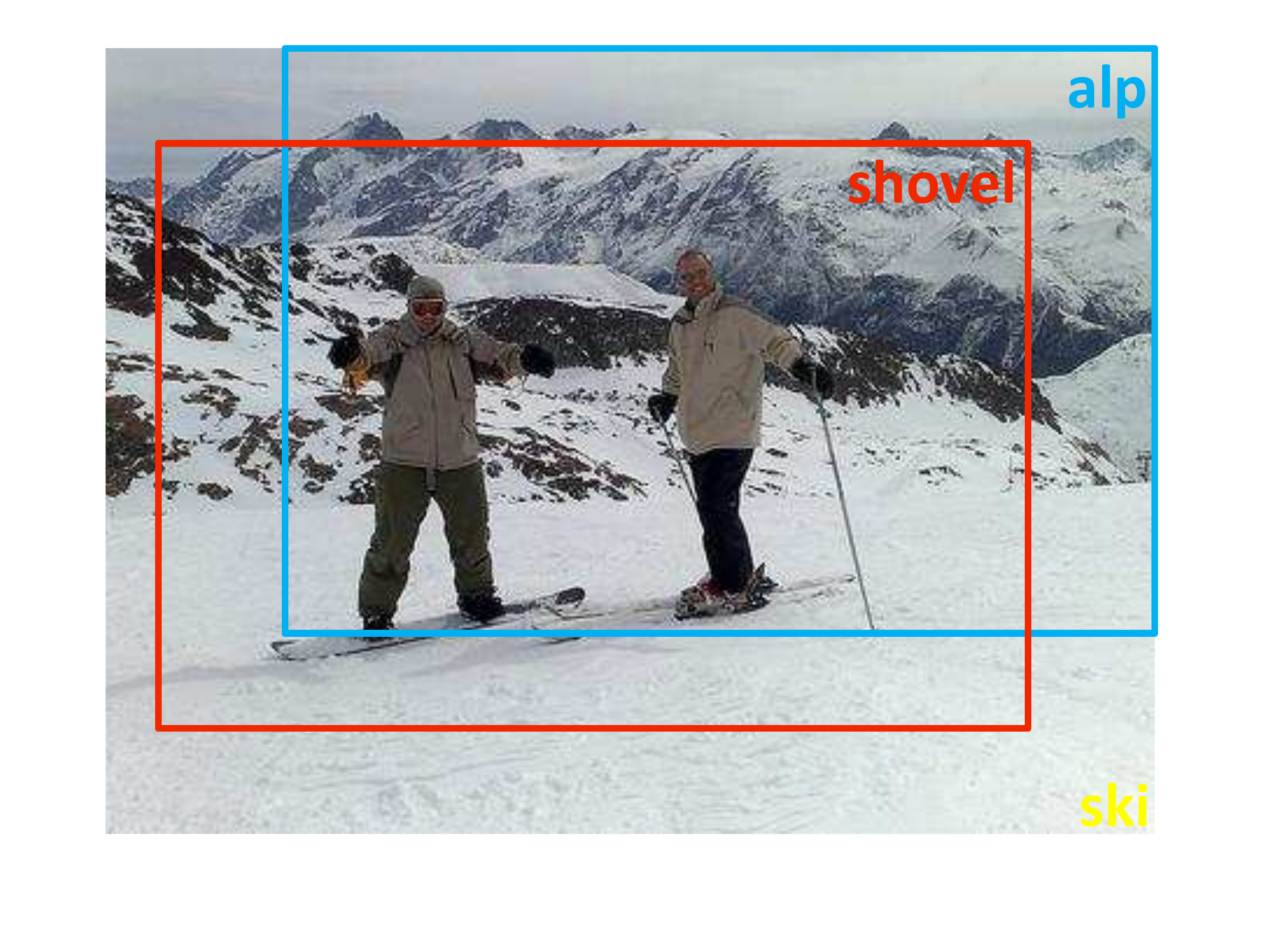}}
\subfigure[bighorn sheep]{\includegraphics[height=0.75in, trim= 0mm 10mm 0mm 15mm]{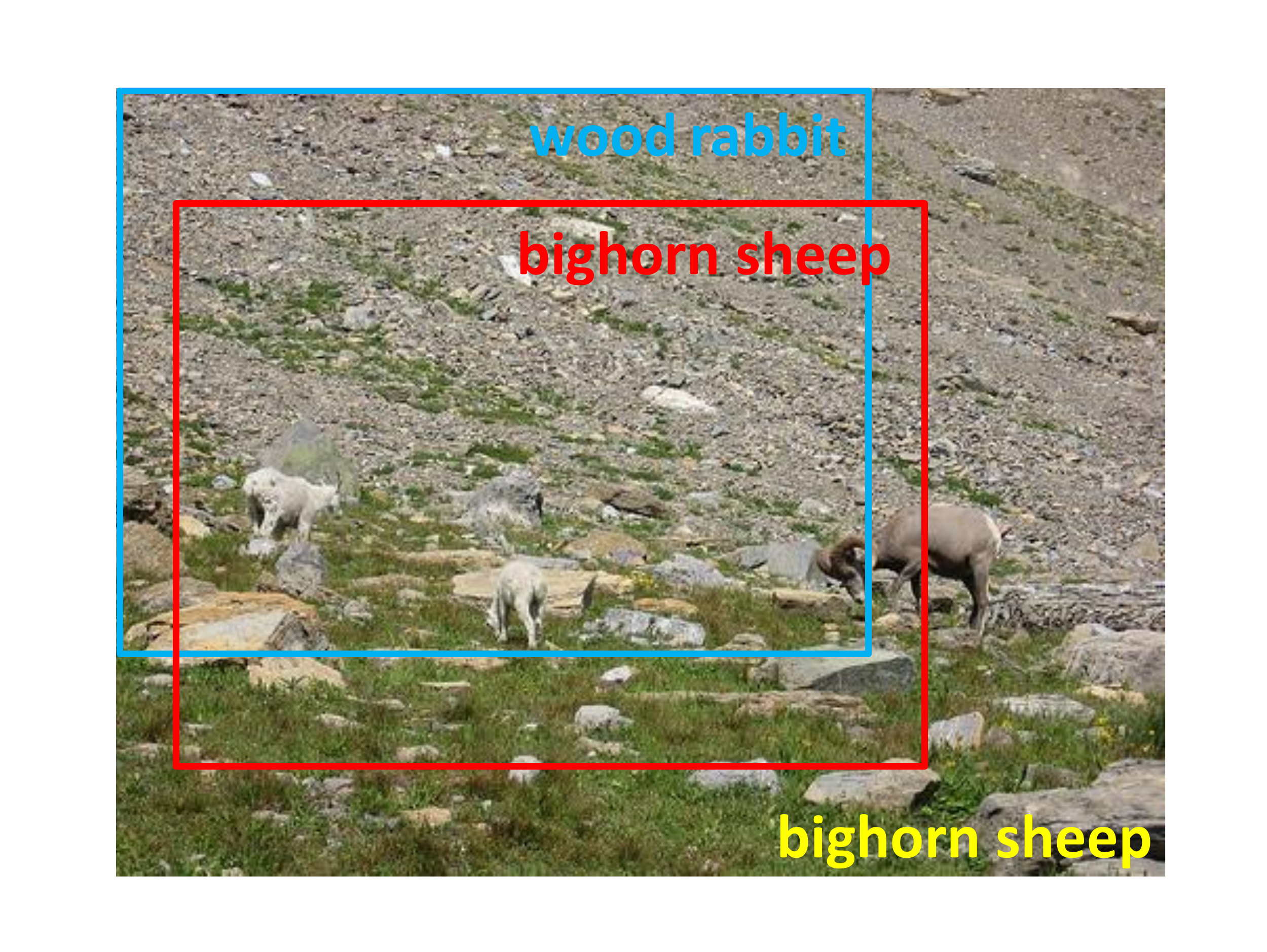}}
\subfigure[pitcher]{\includegraphics[height=0.75in, trim= 0mm 10mm 0mm 15mm]{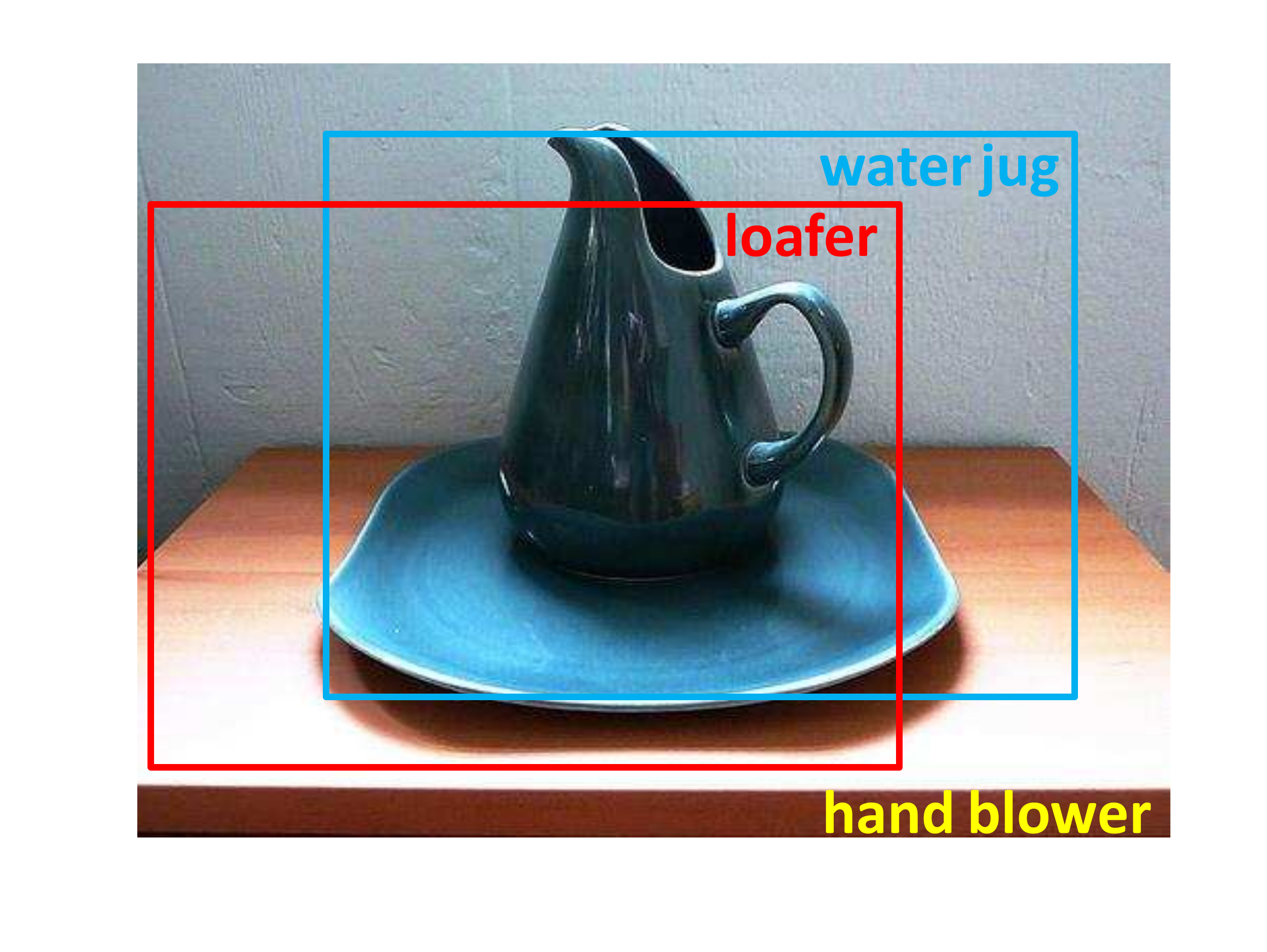}}\\
\subfigure[schooner]{\includegraphics[height=0.75in, trim= 0mm 10mm 0mm 15mm]{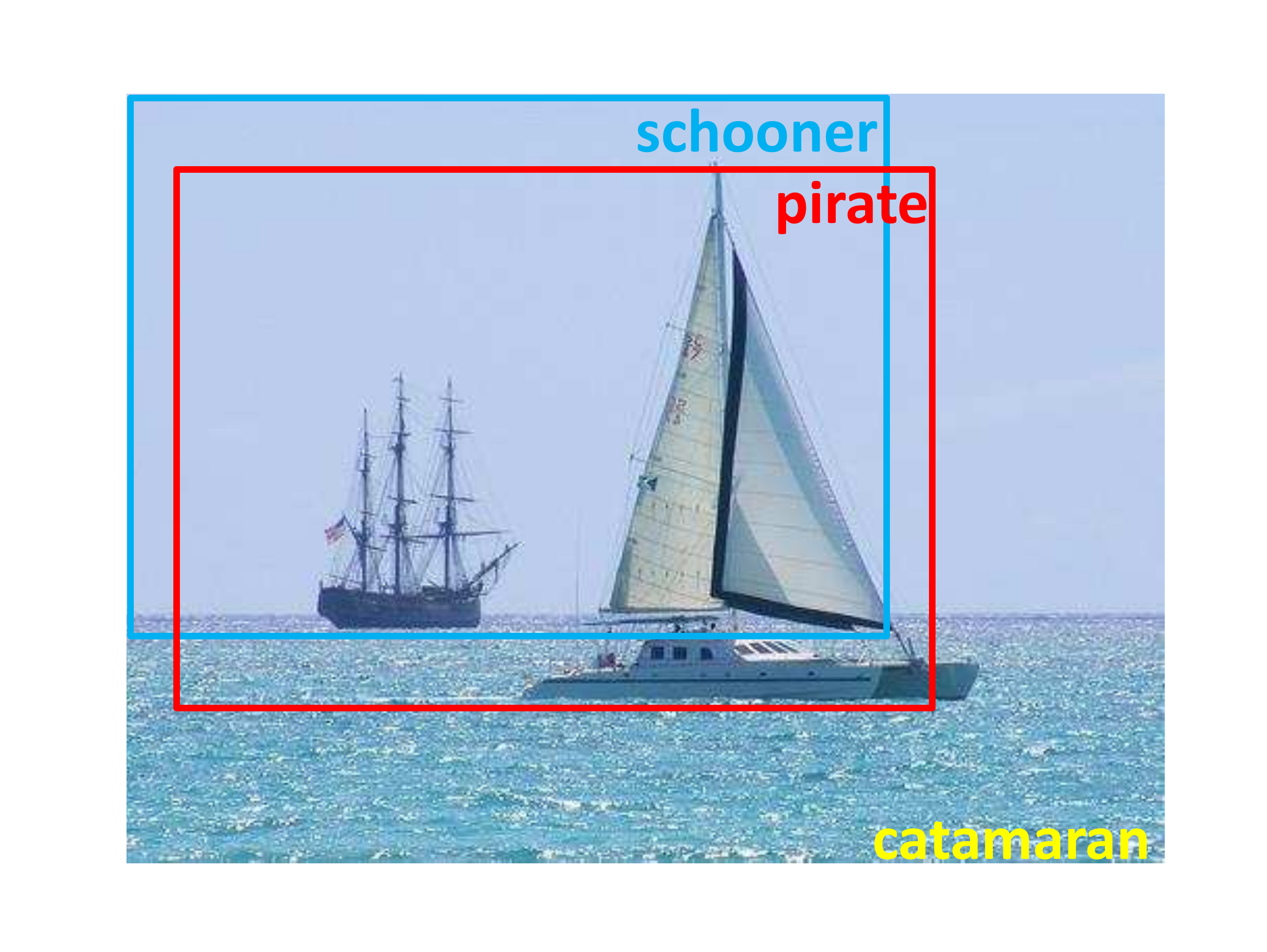}}
\subfigure[bee eater]{\includegraphics[height=0.75in, trim= 0mm 10mm 0mm 15mm]{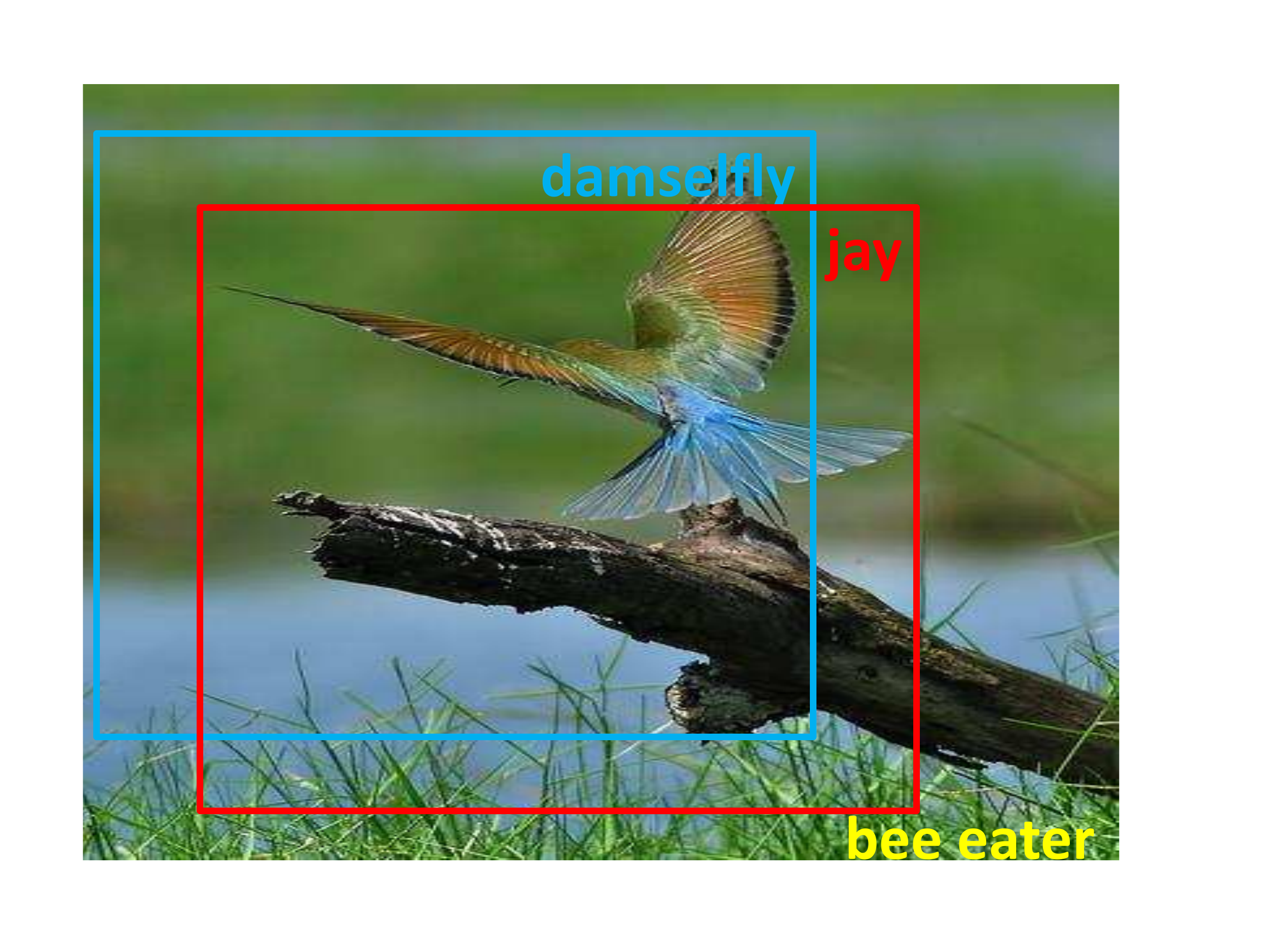}}
\subfigure[alp]{\includegraphics[height=0.75in, trim= 0mm 10mm 0mm 15mm]{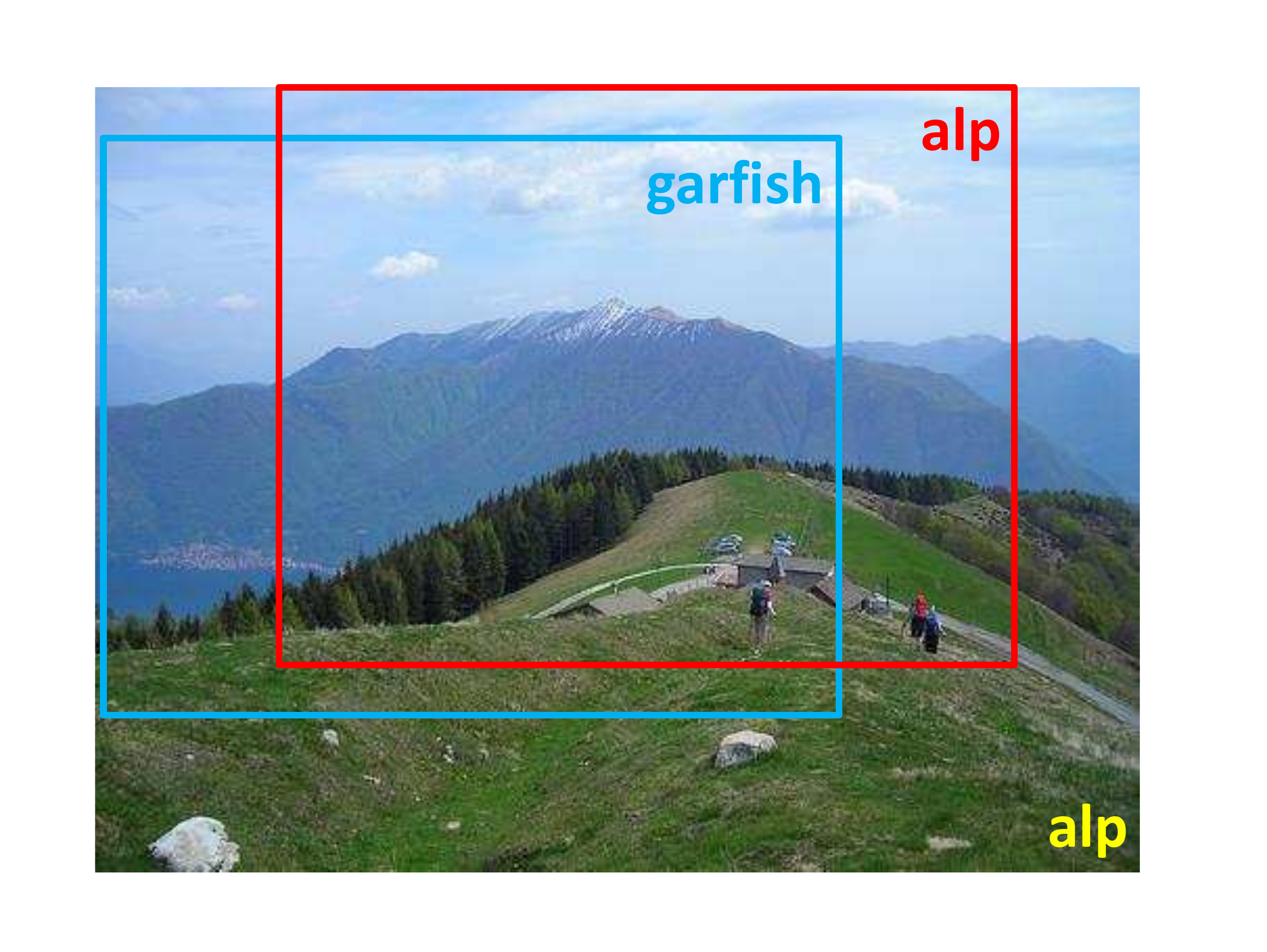}}
\end{tabular} \vspace{-5mm}
\caption{Classification of CNN activations of local patches in an image. The ground truth labels are listed below each image. Labels predicted by whole-image CNN are listed in the bottom right corner.}
\label{fig:variance}
\end{center}
\end{figure*}

\begin{figure*}[!t]
\centering
  \includegraphics[width= 0.8\linewidth]{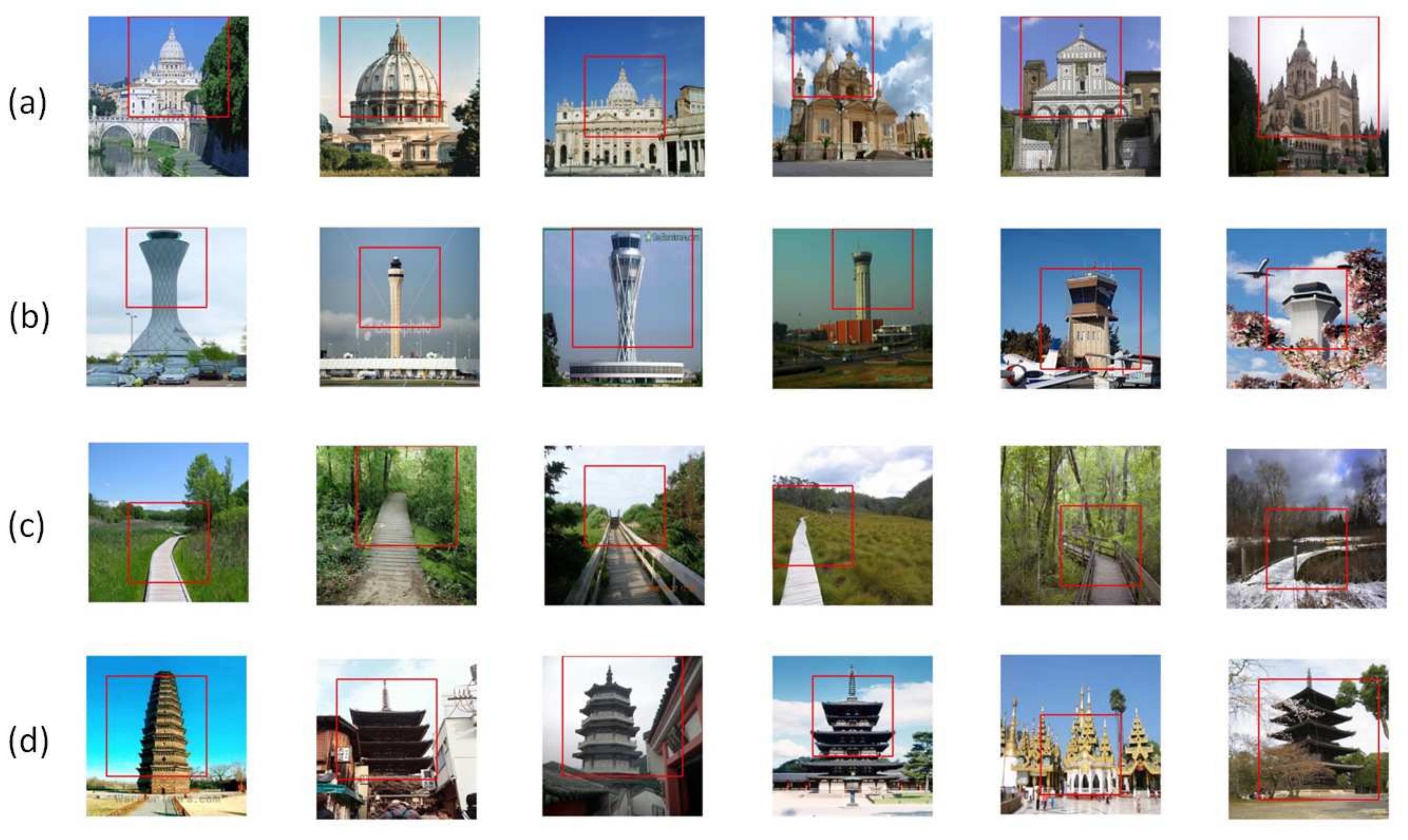} \vspace{-4mm}
    \caption{Highest-response windows (in red) for (a) basilica, (b) control tower, (c) boardwalk, and (d) tower. For each test image resampled to $256\times256$, we search over windows with widths 224, 192, 160, and 128 and a stride of 16 pixels and display the window that gives the highest prediction score for the ground truth category. The detected windows contain similar structures: in (a), (b) and (d), the top parts of towers have been selected; in (c), the windows are all centered on the narrow walkway.}\label{fig:subscene}
\end{figure*}

Figure \ref{fig:variance} further illustrates the lack of robustness of global CNN activations by showing the predictions for a few ILSVRC2012/2013 images based on different image sub-windows. Even for sub-windows that are small translations of each other, the predicted labels can be drastically different. For example, in (f), the red rectangle is correctly labeled ``alp,'' while the overlapping rectangle is incorrectly labeled ``garfish.'' But, while picking the wrong window can give a bad prediction, picking the ``right'' one can give a good prediction: in (d), the whole image is wrongly labeled, but one of its sub-windows can get the correct label -- ``schooner.'' This immediately suggests a sliding window protocol at test time: given a test image, extract windows at multiple scales and locations, compute their CNN activations and prediction scores, and look for the window that gives the maximum score for a given class. Figure \ref{fig:subscene} illustrates such a ``scene detection'' approach~\cite{pandey2011scene,xiao2010sun} on a few SUN images.
In fact, it is already common for CNN implementations to sample multiple windows at test time: the systems of~\cite{krizhevsky2012imagenet,donahue2013decaf,Jia13caffe} can take five sub-image windows corresponding to the center and four corners, together with their flipped versions, and average the prediction scores over these ten windows. As will be shown in Table \ref{imagenet}, for Caffe, this ``center+corner+flip'' strategy gets 56.30\% classification accuracy on ILSVRC2012/2013 vs. 54.34\% for simply classifying global image windows. An even more recent system, OverFeat~\cite{sermanet2013overfeat}, incorporates a more comprehensive multi-scale voting scheme for classification, where efficient computations are used to extract class-level activations at a denser sampling of locations and scales, and the average or maximum of these activations is taken to produce the final classification results. With this scheme, OverFeat can achieve as high as 64.26\% accuracy on ILSVRC2012/2013, albeit starting from a better baseline CNN with 60.72\% accuracy. 

While the above window sampling schemes do improve the robustness of prediction over single global CNN activations, they all combine activations (classifier responses) from the final prediction layer, which means that they can only be used following training (or fine-tuning) for a particular prediction task, and do not naturally produce feature vectors for other datasets or tasks. By contrast, MOP-CNN combines activations of the last fully connected layer, so it is a more generic representation that can even work for tasks like image retrieval, which may be done in an unsupervised fashion and for which labeled training data may not be available. 



\section{Large-Scale Evaluation} \label{experiments}

\begin{table}[t]
\begin{center}
\small{
\caption{\small A summary of baselines and their relationship to the MOP-CNN method. \label{relation}} \vspace{-2mm}
\begin{tabular}{c||cccccc}
\hline
 pooling method  / scale 	&   ~~~multi-scale~~~       &   ~~~concatenation~~~                    \\ \hline
~~~Average pooling~~~  &           Avg (multi-scale)        &          Avg (concatenation)                   \\
~~~Max pooling~~~ &     Max (multi-scale)        &    Max (concatenation)                       \\
~~~VLAD pooling~~~  &    VLAD (multi-scale)        &    \textbf{MOP-CNN}                      \\
 \hline
\end{tabular}
}
\end{center}
\end{table}

\subsection{Baselines} \label{sec:baselines}

 To validate MOP-CNN, we need to demonstrate that a simpler patch sampling and pooling scheme cannot achieve the same performance. As simpler alternatives to VLAD pooling, we consider {\bf average pooling}, which involves computing the mean of the 4096-dimensional activations at each scale level, and {\bf maximum pooling}, which involves computing their element-wise maximum. We did not consider standard BoF pooling because it has been demonstrated to be less accurate than VLAD~\cite{jegou2010aggregating}; to get competitive performance, we would need a codebook size much larger than 100, which would make the quantization step prohibitively expensive. As additional baselines, we need to examine alternative strategies with regards to pooling across scale levels. The {\bf multi-scale} strategy corresponds to taking the union of all the patches from an image, regardless of scale, and pooling them together. The {\bf concatenation} strategy refers to pooling patches from three levels separately and then concatenating the result. Finally, we separately examine the performance of individual scale levels as well as concatenations of just pairs of them. In particular, {\bf level1} is simply the 4096-dimensional global descriptor of the entire image, which was suggested in \cite{donahue2013decaf} as a generic image descriptor. These baselines and their relationship to our full MOP-CNN scheme are summarized in Table \ref{relation}.

\subsection{Datasets}

We test our approach on four well-known benchmark datasets:\smallskip

\noindent{\textbf{SUN397}}~\cite{xiao2010sun} is the largest dataset to date for scene recognition. It contains 397 scene categories and each has at least 100 images. The evaluation protocol involves training and testing on ten different splits and reporting the average classification accuracy. The splits are fixed and publicly available from~\cite{xiao2010sun}; each has 50 training and 50 test images.  \smallskip

\noindent{\textbf{MIT Indoor}}~\cite{Quattoni09} contains 67 categories. While outdoor scenes, which comprise more than half of SUN (220 out of 397), can often be characterized by global scene statistics, indoor scenes tend to be much more variable in terms of composition and better characterized by the objects they contain. This makes the MIT Indoor dataset an interesting test case for our representation, which is designed to focus more on appearance of sub-image windows and have more invariance to global transformations. The standard training/test split for the Indoor dataset consists of 80 training and 20 test images per class. \smallskip

\noindent{\textbf{ILSVRC2012/2013}}~\cite{deng12,Russakovsky13}, or ImageNet Large-Scale Visual Recognition Challenge, is the most prominent benchmark for comparing large-scale image classification methods and is the dataset on which the Caffe representation we use~\cite{Jia13caffe} is pre-trained. ILSVRC differs from the previous two datasets in that most of its categories focus on objects, not scenes, and the objects tend to be highly salient and centered in images. It contains 1000 classes corresponding to leaf nodes in ImageNet. Each class has more than 1000 unique training images, and there is a separate validation set with 50,000 images. The 2012 and 2013 versions of the ILSVRC competition have the same training and validation data. Classification accuracy on the validation set is used to evaluate different methods.  \smallskip

\noindent{\textbf{INRIA Holidays}}~\cite{Jegou08} is a standard benchmark for image retrieval. It contains 1491 images corresponding to 500 image instances. Each instance has 2-3 images describing the same object or location. A set of 500 images are used as queries, and the rest are used as the database. Mean average precision (mAP) is the evaluation metric.

\subsection{Image Classification Results}

In all of the following experiments, we train classifiers using the linear SVM implementation from the INRIA JSGD package~\cite{akata2013good}. We fix the regularization parameter to $10^{-5}$ and the learning rate to $0.2$, and train for 100 epochs.

\begin{table}[t]
\begin{center}
\small{
\caption{\small Scene recognition on SUN397. (a)  Alternative pooling baselines (see Section \ref{sec:baselines} and Table \ref{relation});
(b) Different combinations of scale levels -- in particular, ``level1'' corresponds to the global CNN representation and ``level1+level2+level3''
corresponds to the proposed MOP-CNN method.
(c) Published numbers for state-of-the-art methods.
\label{sun}} \vspace{-4mm}
\begin{tabular}{lc||c|ccccc}
\hline
& method   	&  ~~feature dimension~~         &        ~~accuracy~~                \\ \hline
(a) &   Avg  (Multi-Scale)  &     4,096         &            39.62                               \\
&     Avg  (Concatenation)  &     12,288         &      47.50                             \\
&   Max  (Multi-Scale)  &     4,096         &       43.51                                    \\
&  Max  (Concatenation)  &     12,288         &    48.50                               \\
&     VLAD  (Multi-Scale)  &    4,096        &           47.32                              \\
   \hline
(b) &   level1  &     4,096         &         39.57                                \\
&   level2  &       4,096      &          45.34                              \\
&   level3  &      4,096       &          40.21                                   \\
&   level1 + level2  &    8,192        &  49.91                                       \\
&   level1 + level3  &    8,192           &   49.52                                    \\
&   level2 + level3  &    8,192           &   49.66                                    \\
&   level1 + level2 + level3 (MOP-CNN) &   12,288              &          \textbf{51.98}                \\   \hline
(c) &   Xiao et al. \cite{xiao2010sun}    &     --          &        38.00                    \\
&   DeCAF \cite{donahue2013decaf}    &     4,096         &         40.94                    \\
& FV (SIFT + Local Color Statistic) \cite{sanchez:hal-00830491}   &     256,000         &         47.20                    \\ \hline
\end{tabular}}
\end{center}
\end{table}

\begin{figure}[!t]
\hspace{0.5cm}
  \includegraphics[width=5in, trim= 12mm 10mm -20mm 5mm]{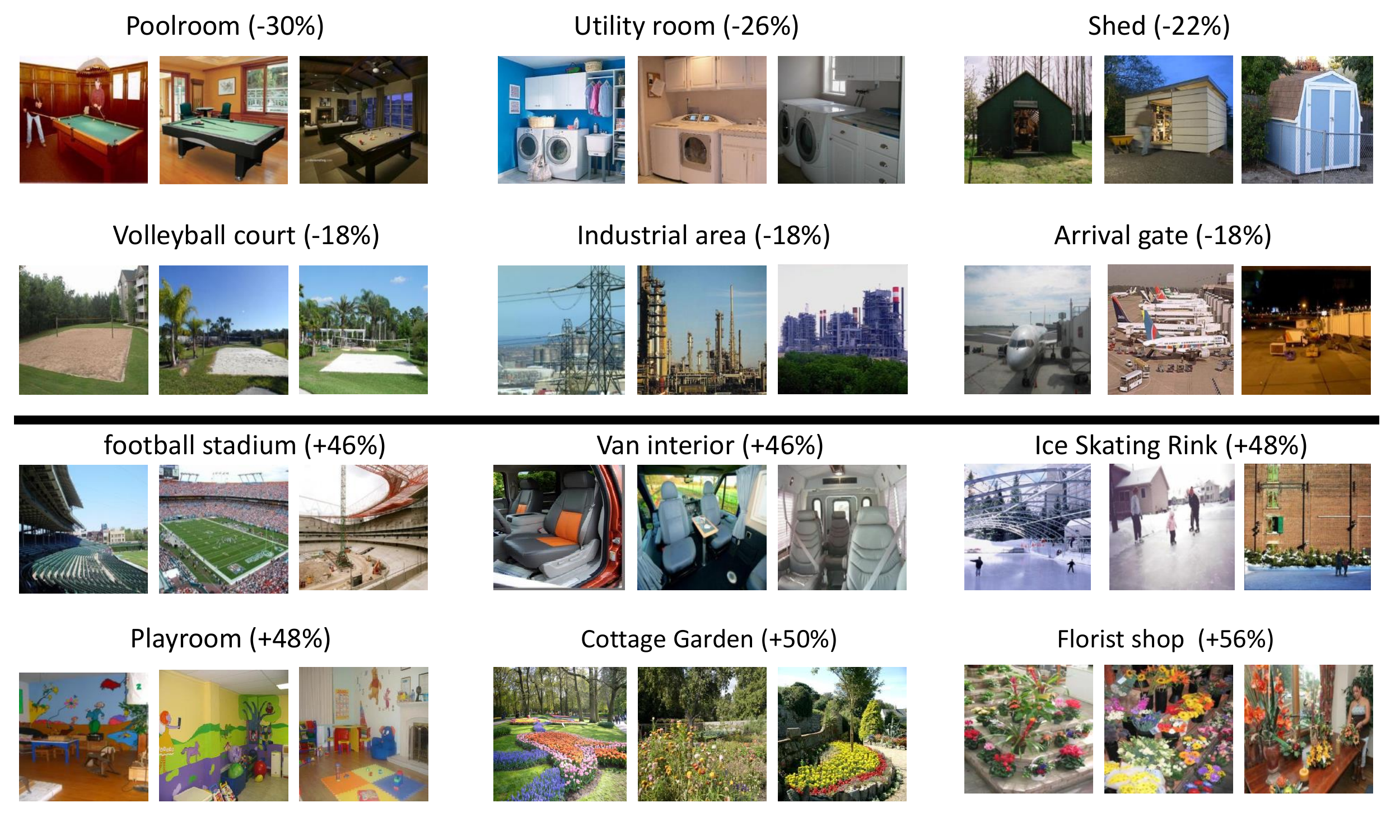}
    \caption{SUN classes on which MOP-CNN gives the biggest decrease over the level1 global features (top), and classes on which it gives the biggest increase (bottom).}\label{fig:sun_correct}\vspace{5mm}
\end{figure}

Table \ref{sun} reports our results on the SUN397 dataset. From the results for baseline pooling methods in (a), we can see that VLAD works better than average and max pooling and that pooling scale levels separately works better than pooling them together (which is not altogether surprising, since the latter strategy raises the feature dimensionality by a factor of three). From (b), we can see that concatenating all three scale levels gives a significant improvement over any subset.
For reference, Part (c) of Table \ref{sun} gives published state-of-the-art results from the literature. Xiao et al.~\cite{xiao2010sun}, who have collected the SUN dataset, have also published a baseline accuracy of 38\% using a combination of standard features like GIST, color histograms, and BoF. This baseline is slightly exceeded by the level1 method, i.e., global 4096-dimensional Caffe activations pre-trained on ImageNet. The Caffe accuracy of 39.57\% is also comparable to the 40.94\% with an analogous setup for DeCAF~\cite{donahue2013decaf}.\footnote{DeCAF is an earlier implementation from the same research group and Caffe is its ``little brother.'' The two implementations are similar, but Caffe is faster, includes support for both CPU and GPU, and is easier to modify.} However, these numbers are still worse than the 47.2\% achieved by high-dimensional Fisher Vectors~\cite{sanchez:hal-00830491} -- to our knowledge, the state of the art on this dataset to date. With our MOP-CNN pooling scheme, we are able to achieve 51.98\% accuracy with feature dimensionality that is an order of magnitude lower than that of~\cite{sanchez:hal-00830491}. Figure \ref{fig:sun_correct} shows six classes on which MOP-CNN gives the biggest improvement over level1, and six on which it has the biggest drop. For classes having an object in the center, MOP-CNN usually cannot improve too much, or might hurt performance. However, for classes that have high spatial variability, or do not have a clear focal object, it can give a substantial improvement.

\begin{table}[t]
\begin{center}
\small{
\caption{\small Classification results on MIT Indoor Scenes.
(a) Alternative pooling baselines (see Section \ref{sec:baselines} and Table \ref{relation});
(b) Different combinations of scale levels;
(c) Published numbers for state-of-the-art methods.
\label{indoor}} \vspace{-2mm}
\begin{tabular}{lc||c|ccc}
\hline
& method   	&   feature dimension         &        ~~accuracy~~          \\ \hline
(a) &   Avg  (Multi-Scale)  &     4,096         &       56.72                                    \\
&   Avg  (Concatenation)  &     12,288         &      65.60                                 \\
&   Max  (Multi-Scale)  &     4,096         &       60.52                                    \\
&   Max  (Concatenation)  &    12,288        &      64.85                                   \\
&   VLAD  (Multi-Scale)  &    4,096        &         66.12                                \\
  \hline
(b) &   level1  &       4,096      &        53.73            \\
&   level2  &       4,096      &        65.52               \\
&   level3  &       4,096      &        62.24                              \\
&   level1 + level2  &    8,192             &           66.64             \\
&   level1 + level3  &     8,192             &          66.87              \\
&   level2 + level3  &    8,192           &           67.24                            \\
&   level1 + level2 + level3 (MOP-CNN)  &  12,288       &        \textbf{68.88}                        \\   \hline
(c) &  SPM  \cite{lazebnik06}   &        5,000    &       34.40             \\
& Discriminative patches  \cite{Singh2012DiscPat}   &   --       &       38.10               \\
&  Disc. patches+GIST+DPM+SPM  \cite{Singh2012DiscPat}   &   --          &         49.40          \\
& FV + Bag of parts~\cite{Juneja13}  &    221,550        &    63.18       \\
&  Mid-level elements~\cite{doersch2013mid}    &       60,000       &        64.03          \\   \hline
\end{tabular}}
\end{center}
\end{table}

\begin{table}[t]
\begin{center}
\small{
\caption{\small Classification results on ILSVRC2012/2013.
(a) Alternative pooling baselines (see Section \ref{sec:baselines} and Table \ref{relation});
(b) Different combinations of scale levels;
(c) Numbers for state-of-the-art CNN implementations. All the numbers come from the respective papers, except the Caffe numbers,
which were obtained by us by directly testing their full network pre-trained on ImageNet. ``Global'' corresponds to
testing on global image features, and ``Center+Corner+Flip'' corresponds to averaging the prediction scores over ten crops taken from
the test image (see Section \ref{s3} for details).
 \label{imagenet}} \vspace{-2mm}
\begin{tabular}{lc||c|ccc}
\hline
& method   	&     feature dimension             &     ~~accuracy~~           \\ \hline
(c) &   Avg  (Multi-Scale)  &     4096         &    53.34                                       \\
&    Avg  (Concatenation)  &    12,288        &  56.12                                 \\
&   Max  (Multi-Scale)  &     4096         &   54.37                                        \\
&     Max  (Concatenation)  &   12,288        &    55.88                               \\
&        VLAD  (Multi-Scale)  &    4,096        &         48.54                                \\
  \hline
(b) &   level1  &     4,096      &        51.46              \\
&   level2  &    4,096         &        48.21          \\
&   level3  &     4,096           &    38.20                         \\
&   level1 + level2  &   8,192         &           56.82                       \\
&   level1 + level3  &    8,192             &      55.91                            \\
&   level2 + level3  &    8,192           &           51.52                            \\
&   level1 + level2 + level3 (MOP-CNN)  &  12,288              &       57.93           \\   \hline
(c) &   Caffe (Global) \cite{Jia13caffe}     &     --          &     54.34              \\
&     Caffe (Center+Corner+Flip) \cite{Jia13caffe}     &     --          &    56.30               \\
& Krizhevsky et al.~\cite{krizhevsky2012imagenet}     &  --     &       {59.93}    &         &             \\
& Zeiler and Fergus (6 CNN models)~\cite{zeiler2013visualizing} & -- & 64.00 \\
&   OverFeat (1 CNN model)  \cite{sermanet2013overfeat}      &     --          &   {\bf 64.26}               \\
&   OverFeat (7 CNN models)  \cite{sermanet2013overfeat}      &     --          &   {\bf 66.04}               \\
   \hline
\end{tabular}}
\end{center}
\end{table}

Table \ref{indoor} reports results on the MIT Indoor dataset. Overall, the trends are consistent with those on SUN, in that VLAD pooling outperforms average and max pooling and combining all three levels yields the best performance. There is one interesting difference from Table \ref{sun}, though: namely, level2 and level3 features work much better than level1 on the Indoor dataset, whereas the difference was much less pronounced on SUN. This is probably because indoor scenes are better described by local patches that have highly distinctive appearance but can vary greatly in terms of location. In fact, several recent methods achieving state-of-the-art results on this dataset are based on the idea of finding such patches~\cite{doersch2013mid,Juneja13,Singh2012DiscPat}. Our MOP-CNN scheme outperforms all of them -- 68.88\% vs. 64.03\% for the method of Doersch et al.~\cite{doersch2013mid}.

Table \ref{imagenet} reports results on ILSVRC2012/2013. The trends for alternative pooling methods in (a) are the same as before. Interestingly, in (b) we can see that, unlike on SUN and MIT Indoor, level2 and level3 features do not work as well as level1. This is likely because the level1 feature was specifically trained on ILSVRC, and this dataset has limited geometric variability. Nevertheless, by combining the three levels, we still get a significant improvement. 
Note that directly running the full pre-trained Caffe network on the global features from the validation set gives 54.34\% accuracy (part (c) of Table \ref{imagenet}, first line), which is higher than our level1 accuracy of 51.46\%. The only difference between these two setups, ``Caffe (Global)'' and ``level1,'' are the parameters of the last classifier layer -- i.e., softmax and SVM, respectively. For Caffe, the softmax layer is jointly trained with all the previous network layers using multiple random windows cropped from training images, while our SVMs are trained separately using only the global image features. Nevertheless, the accuracy of our final MOP-CNN representation, at 57.93\%, is higher than that of the full pre-trained Caffe CNN tested either on the global features (``Global'') or on ten sub-windows (``Center+Corner+Flip''). 

It is important to note that in absolute terms, we do not achieve state-of-the-art results on ILSVRC.
For the 2012 version of the contest, the highest results were achieved by Krizhevsky et al. \cite{krizhevsky2012imagenet}, who have reported a top-1 classification accuracy of 59.93\%. Subsequently, Zeiler and Fergus~\cite{zeiler2013visualizing} have obtained 64\% by refining the Krizhevsky architecture and combining six different models. For the 2013 competition, the highest reported top-1 accuracies are those of Sermanet et al.~\cite{sermanet2013overfeat}: they obtained 64.26\% by aggregating CNN predictions over multiple sub-window locations and scales (as discussed in Section \ref{s3}), and 66.04\% by combining seven such models. While our numbers are clearly lower, it is mainly because our representation is built on Caffe, whose baseline accuracy is below that of~\cite{krizhevsky2012imagenet,zeiler2013visualizing,sermanet2013overfeat}. We believe that MOP-CNN can obtain much better performance when combined with these better CNN models, or by combining multiple independently trained CNNs as in~\cite{zeiler2013visualizing,sermanet2013overfeat}.


\subsection{Image Retrieval Results}

\begin{table}[!t]
\begin{center}
\small{
\caption{\small Image retrieval results on the Holidays dataset.
(a) Alternative pooling baselines (see Section \ref{sec:baselines} and Table \ref{relation});
(b) Different combinations of scale levels;
(c) Full MOP-CNN descriptor vector compressed by PCA and followed by whitening~\cite{jegou:hal-00722622}, for two different output dimensionalities;
(c) Published state-of-the-art results with a compact global descriptor (see text for discussion).
\label{holiday}} \vspace{-2mm}
\begin{tabular}{lc||c|ccc}
\hline
& method   	&  ~~feature dimension~~  &  ~~mAP~~          \\ \hline
(a) &   Avg  (Multi-Scale)  &     4,096         &       71.32                                    \\
&     Avg  (Concatenation)  &     12,288         &     75.02                              \\
&   Max  (Multi-Scale)  &     4,096         &       76.23                                    \\
&     Max  (Concatenation)  &     12,288       &      75.07                             \\
&        VLAD  (Multi-Scale)  &    4,096        &     78.42                                    \\
  \hline
(b) &   level1  &   4,096     &        70.53            \\
&   level2  &   4,096       &     74.02        \\
&   level3  &   4,096     &       75.45                      \\
&   level1 + level2  &   8,192   &     75.86                             \\
&   level1 + level3  &    8,192 &     78.92                                 \\
&      level2 + level3  &    8,192           &     77.91                                  \\
&   level1 + level2 + level3 (MOP-CNN)  &  12,288   &   78.82                              \\   \hline
(c) &  MOP-CNN + PCA + Whitening &   512   &         78.38                   \\
&  MOP-CNN + PCA + Whitening  &   2048   &     \textbf{80.18}                    \\   \hline
(d) &      FV  \cite{jegou2010aggregating}   &  8,192      &   62.50      &             \\
& FV + PCA  \cite{jegou2010aggregating}  &    256    &    62.60     &             \\
& Gordo et al.  \cite{gordo2012leveraging}  &    512    &    78.90     &             \\
  \hline
\end{tabular}} 
\end{center} 
\end{table}

\begin{figure}[!t]
\begin{center}
\footnotesize{
\begin{tabular}{ccc}
      \includegraphics[width=0.5in, clip=true, trim= 0mm -28mm 0mm 0mm]{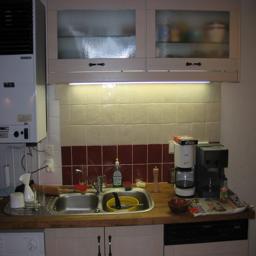} &
   \includegraphics[width=1.62in, clip=true, trim= 15mm 100mm 25mm 90mm]{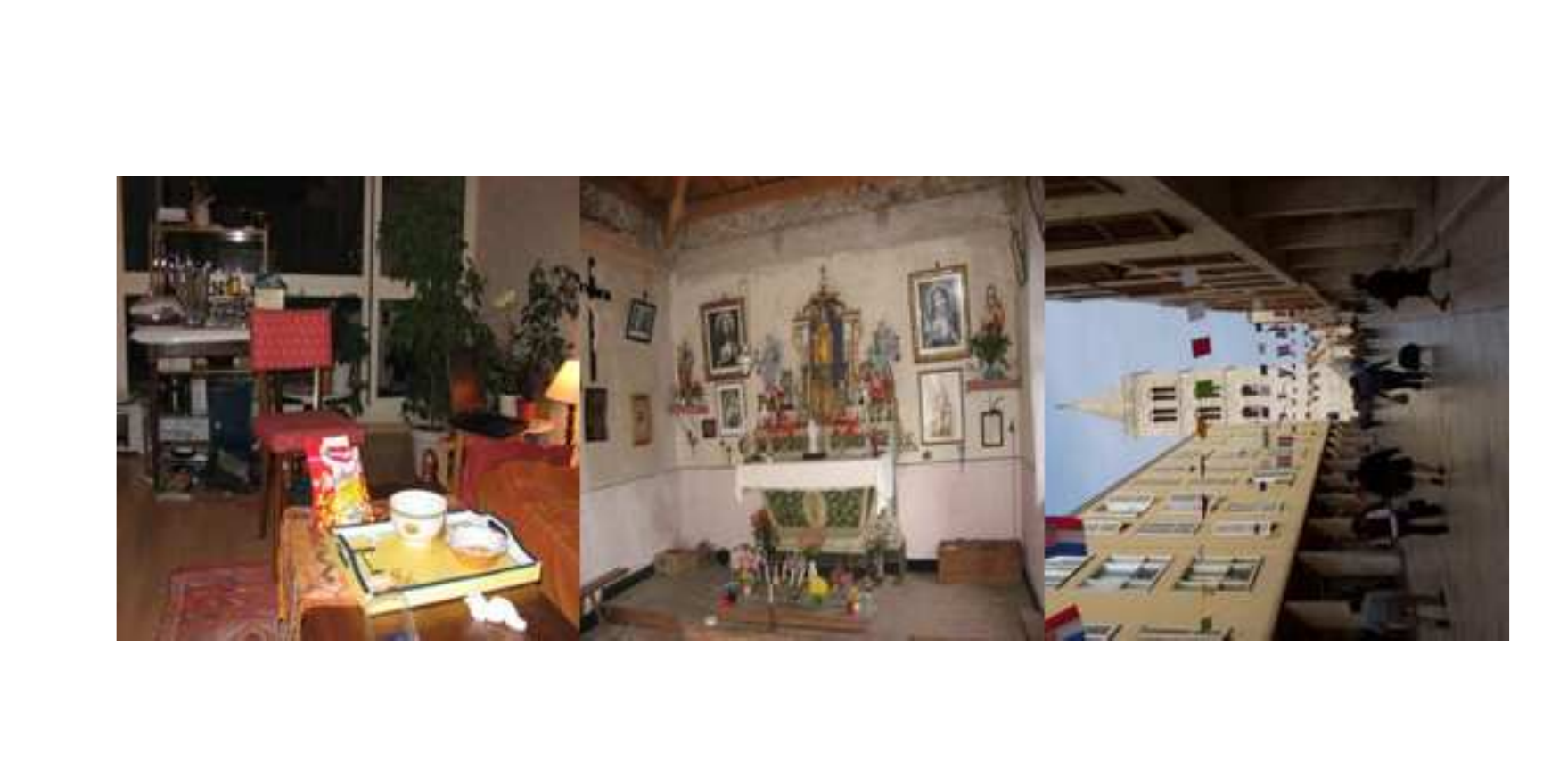} &
   \includegraphics[width=1.62in,  trim= 20mm 100mm 20mm 90mm]{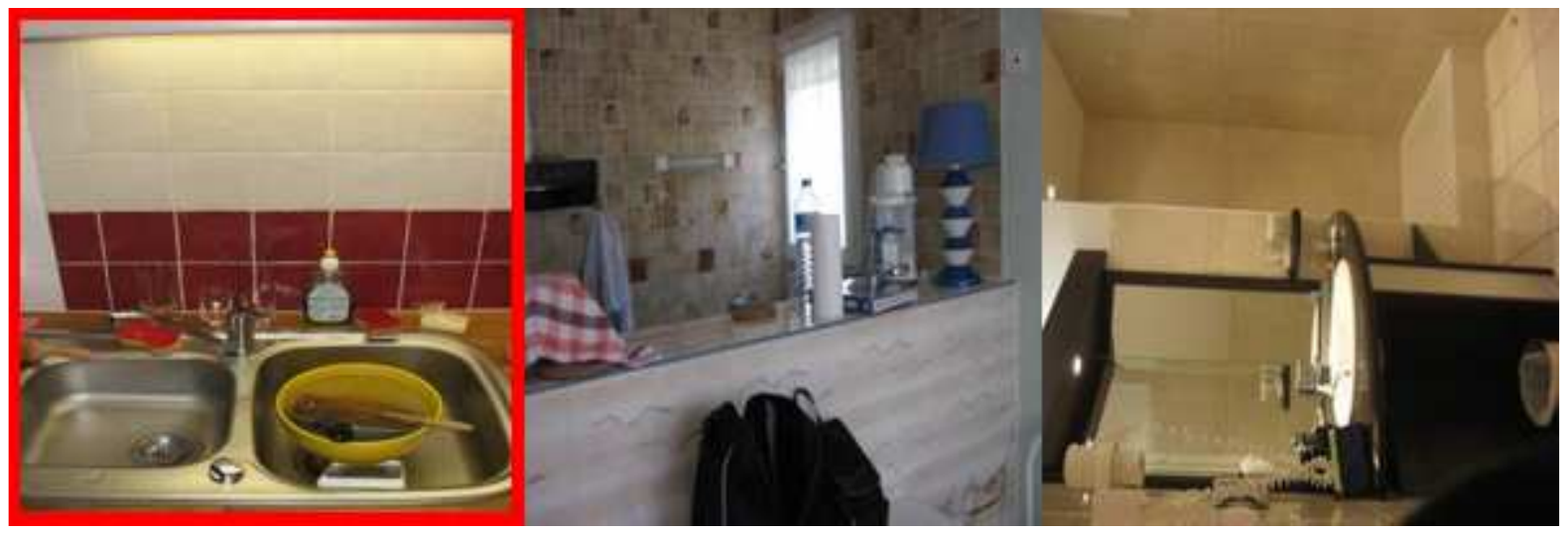}\\
      \includegraphics[width=0.5in,  trim= 0mm -28mm 0mm 10mm]{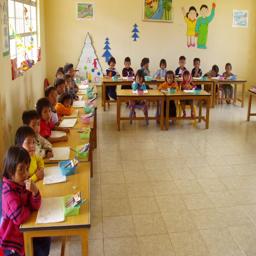} &
   \includegraphics[width=1.62in,  trim= 15mm 100mm 25mm 140mm]{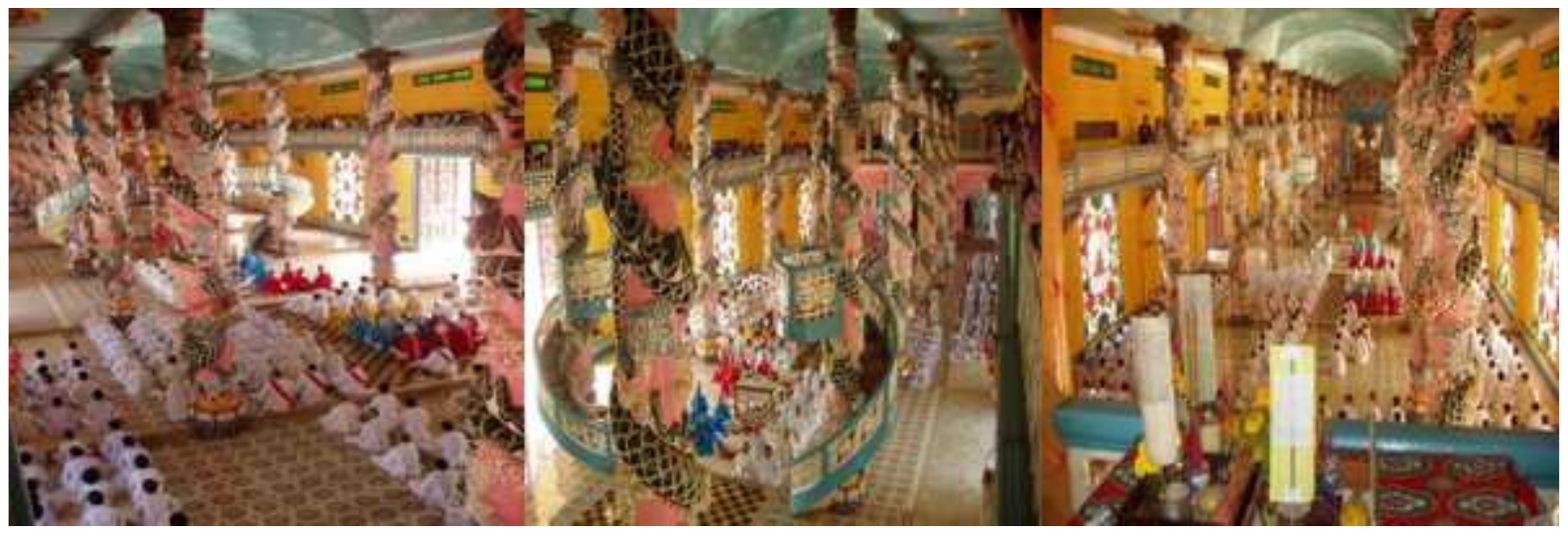} &
   \includegraphics[width=1.62in,  trim= 20mm 100mm 20mm 140mm]{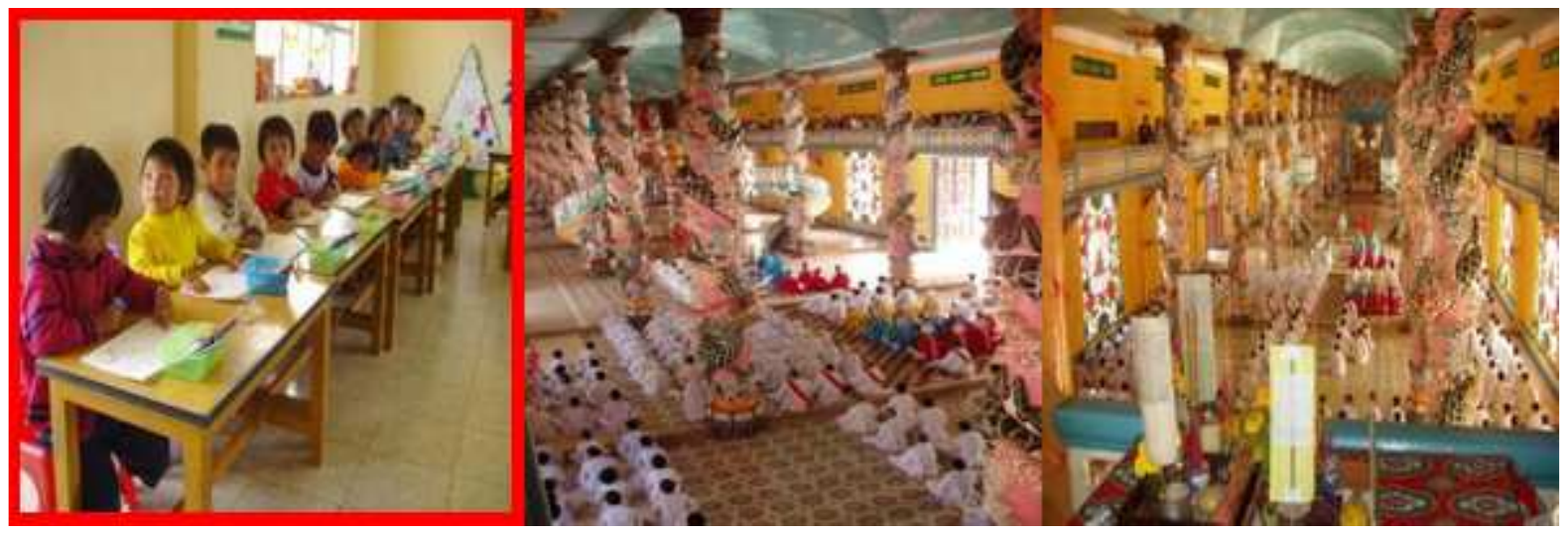} \\
      \includegraphics[width=0.5in,  trim= 0mm -28mm 0mm 15mm]{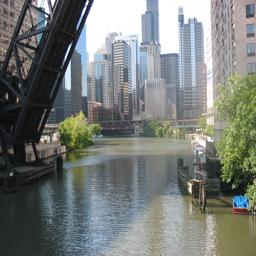} &
   \includegraphics[width=1.62in,  trim= 15mm 100mm 25mm 145mm]{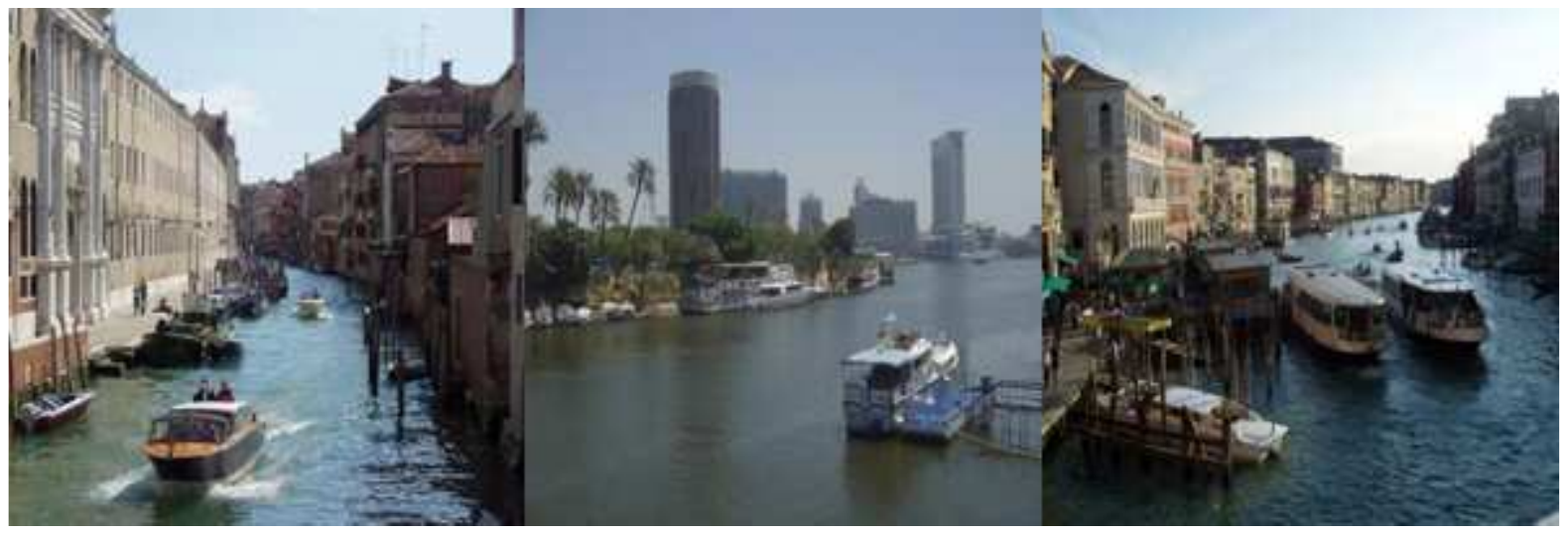} &
   \includegraphics[width=1.62in,  trim= 20mm 100mm 20mm 145mm]{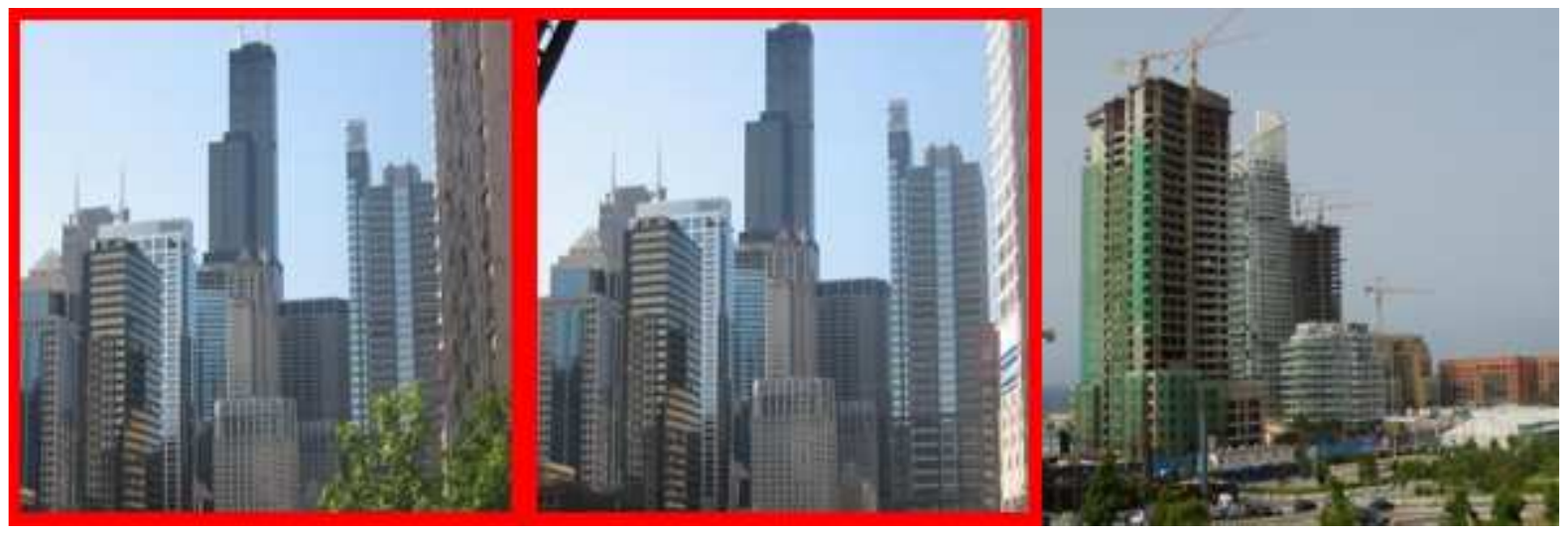} \vspace{-4mm} \\
(a) Query & (b) level1 & (c) MOP-CNN
\end{tabular}}
\end{center}
\vspace{-4mm}
   \caption{Image retrieval examples on the Holiday dataset. Red border indicates a ground truth image (i.e., a positive retrieval result). We only show three retrieved examples per query because each query only has one to two ground truth images.}
   \label{holidaysample}
\end{figure}

As our last experiment, we demonstrate the usefulness of our approach for an \emph{unsupervised} image retrieval scenario on the Holidays dataset.
Table \ref{holiday} reports the mAP results for nearest neighbor retrieval of feature vectors using the Euclidean distance. On this dataset, level1 is the weakest of all three levels because images of the same instance may be related by large rotations, viewpoint changes, etc., and global CNN activations do not have strong enough invariance to handle these transformations. As before, combining all three levels achieves the best performance of 78.82\%. Using aggressive dimensionality reduction with PCA and whitening as suggested in~\cite{jegou:hal-00722622}, we can raise the mAP even further to 80.8\% with only a 2048-dimensional feature vector. The state of the art performance on this dataset with a compact descriptor is obtained by Gordo et al.~\cite{gordo2012leveraging} by using FV/VLAD and discriminative dimensionality reduction, while our method still achieves comparable or better performance. Note that it is possible to obtain even higher results on Holidays with methods based on inverted files with very large vocabularies. In particular, Tolias et al.~\cite{Tolias13} report 88\% but their representation would take more than 4 million dimensions per image if expanded into an explicit feature vector, and is not scalable to large datasets. Yet further improvements may be possible by adding techniques such as query expansion and geometric verification, but they are not applicable for generic image representation, which is our main focus. Finally, we show retrieval examples in Figure \ref{holidaysample}. We can clearly see that MOP-CNN has improved robustness to shifts, scaling, and viewpoint changes over global CNN activations.


\section{Discussion} \label{sec:discussion}

This paper has presented a multi-scale orderless pooling scheme that is built on top of deep activation features of local image patches. On four very challenging datasets, we have achieved a substantial improvement over global CNN activations, in some cases outperforming the state of the art. These results are achieved with the same set of parameters (i.e., patch sizes and sampling, codebook size, PCA dimension, etc.), which clearly shows the good generalization ability of the proposed approach. As a generic low-dimensional image representation, it is not restricted to supervised tasks like image classification, but can also be used for unsupervised tasks such as retrieval.

Our work opens several promising avenues for future research. First, it remains interesting to investigate more sophisticated ways to incorporate orderless information in CNN. One possible way is to change the architecture of current deep networks fundamentally to improve their holistic invariance. Second, the feature extraction stage of our current pipeline is somewhat slow, and it is interesting to exploit the convolutional network structure to speed it up. Fortunately, there is fast ongoing progress in optimizing this step. One example is the multi-scale scheme of Sermanet et al.~\cite{sermanet2013overfeat} mentioned earlier, and another is DenseNet~\cite{Iandola14}. In the future, we would like to reimplement MOP-CNN to benefit from such architectures. 
\smallskip

\noindent {\bf Acknowledgments.} Lazebnik's research was partially supported by NSF grants 1228082 and 1302438, the DARPA Computer Science Study Group, Xerox UAC, Microsoft Research, and the Sloan Foundation. Gong was supported by the 2013 Google Ph.D. Fellowship in Machine Perception.

\footnotesize
\bibliographystyle{splncs}
\bibliography{egbib}

\end{document}